\documentclass[11pt]{article}

\usepackage[T1]{fontenc}
\usepackage[utf8]{inputenc}
\usepackage{lmodern}
\usepackage[margin=1in]{geometry}
\usepackage{microtype}

\usepackage{graphicx}
\usepackage{amsmath, amssymb}
\usepackage{booktabs}
\usepackage{csquotes}
\usepackage{caption}
\usepackage{subcaption}
\usepackage{float}
\usepackage{placeins}
\usepackage{xurl}
\usepackage{pdflscape}
\usepackage[hidelinks]{hyperref}

\usepackage[backend=biber,style=apa]{biblatex}
\title{When Should a Language Model Trust Itself?\\
Same-Model Self-Verification as a Conditional Confidence Signal}

\author{
Aditya Ajay Phalod \\
Independent Researcher \\
\texttt{aphalod@student.ubc.ca}
}

\date{}
\begin{document}

\maketitle

\setlength{\textfloatsep}{10pt plus 2pt minus 2pt}
\setlength{\floatsep}{8pt plus 2pt minus 2pt}
\setlength{\intextsep}{8pt plus 2pt minus 2pt}

\begin{abstract}
Same-model self-verification, prompting a model to audit its own predicted answer, is a plausible confidence signal for selective prediction, but its practical value remains unclear once strong likelihood-based baselines are taken seriously. We evaluate self-verification against two such baselines, LL-AVG and LL-SUM, on ARC-Challenge and TruthfulQA-MC across multiple model families, scales, and prompt variants. We measure not only correctness ranking, but also abstention quality through AURC and operating-point analyses. The results are sharply task- and model-dependent. On ARC-Challenge, self-verification substantially improves over LL-AVG for \texttt{Phi-2} and the \texttt{Qwen} models, with the largest gains appearing in \texttt{Qwen-7B}. On TruthfulQA-MC, however, the signal is less reliable: smaller models can become prompt-sensitive, \texttt{DeepSeek-R1-Distill-8B} degrades relative to LL-AVG, and LL-SUM often remains the stronger practical baseline. We therefore do not treat self-verification as a general-purpose uncertainty estimator. In this setting, it is better understood as a conditional confidence signal whose value depends on task type, model family, prompt formulation, and, crucially, the baseline it must beat.
\end{abstract}

\section{Introduction}

As language models are deployed in increasingly consequential settings, it is no longer enough for them to be accurate on average; they must also know when \emph{not} to answer. This is the central goal of \emph{selective prediction}: to rank predictions by confidence so that systems can abstain on high-risk cases while preserving coverage on easier ones. In practice, this requires confidence signals that are not only correlated with correctness, but also robust across tasks, model families, and prompting choices.

A natural candidate is \emph{same-model self-verification}: after producing an answer, the model is prompted to judge whether that answer is correct. This idea is appealing because it is simple, architecture-agnostic, and easy to deploy at inference time. It also fits a broader intuition in language-model reliability: if a model can reason through a problem, perhaps it can also recognize when its own reasoning has failed. Prior work has explored related forms of self-evaluation, verbalized confidence, and $P(\mathrm{True})$-style uncertainty estimation, suggesting that language models can sometimes provide useful introspective signals \parencite{kadavath2022languagemodelsmostlyknow}. Related work also shows that elicited confidence can depend materially on prompt design and extraction method \parencite{xiong2023canllmsexpress,yang2024verbalizedconf}. At the same time, recent work has emphasized that language models can remain confidently wrong, making calibrated uncertainty a central requirement for reliability-sensitive deployment \parencite{kalai2025whylanguagemodelshallucinate}.

What is less clear is whether a same-model self-check remains useful once compared against cheap confidence signals already available from the model's answer distribution. This is a comparative, deployment-facing question: an extra verification pass is only worthwhile if it improves selective prediction relative to strong one-pass baselines. More specifically, prior work establishes that self-evaluation can be informative and that elicitation matters, but it leaves open a narrower practical question: when does same-model self-verification improve abstention behavior once compared directly against both LL-AVG and LL-SUM?

In this paper, we study that question in a controlled multiple-choice setting. We compare same-model self-verification against two likelihood-based baselines: \emph{LL-AVG}, which scores an answer using length-normalized option log-likelihood, and \emph{LL-SUM}, which uses the non-length-normalized sum of option log-probabilities. We evaluate these signals on two benchmarks chosen to stress different failure profiles. \textbf{ARC-Challenge} \parencite{clark2018thinkyouhavesolvedquestion} emphasizes difficult reasoning and knowledge integration, where errors may arise from computational or inferential slips. \textbf{TruthfulQA-MC} \parencite{lin2021truthfulqa}, by contrast, is designed to expose systematic misconceptions and truthfulness failures, where wrong answers may reflect deeper representational problems rather than transient reasoning errors. We do not treat this two-benchmark contrast as a clean causal decomposition of error types; rather, it provides a deliberately contrasting testbed for asking whether the value of self-verification changes across settings.

Across these settings, we evaluate multiple model families and scales, including \texttt{Phi-2}, \texttt{Qwen-1.5B}, \texttt{TinyLlama-1.1B}, \texttt{Qwen-7B}, and \texttt{DeepSeek-R1-Distill-8B}, as well as multiple self-verification prompt variants. We measure not only correctness ranking via AUROC, but also abstention quality via AURC and operating-point analyses. This lets us ask a stricter question than whether self-verification is merely correlated with accuracy: when does it improve abstention behavior relative to strong likelihood baselines, and when does it fail to justify its extra pass?

Our results show a clear but narrow pattern. On ARC-Challenge, self-verification substantially improves over LL-AVG for \texttt{Phi-2} and the \texttt{Qwen} models, with the largest gains appearing in \texttt{Qwen-7B}. On TruthfulQA-MC, however, the signal is far less robust: some smaller models become prompt-sensitive, \texttt{DeepSeek-R1-Distill-8B} is consistently weak relative to LL-AVG, and \texttt{LL-SUM} often remains the stronger practical baseline. The contrast between \texttt{Qwen-7B} and \texttt{DeepSeek-R1-Distill-8B} also argues against a simple story in which more scale or reasoning-oriented training automatically yields better introspective confidence.

Taken together, these results suggest that same-model self-verification should not be treated as a general-purpose uncertainty estimator. In this setting, it is better understood as a conditional confidence signal whose value depends on task regime, model family, prompt formulation, and the baseline to which it is compared.

Our contributions are as follows:
\begin{itemize}
    \item We provide a comparative evaluation of same-model self-verification as a confidence signal for selective prediction, benchmarking it against both length-normalized (\texttt{LL-AVG}) and unnormalized (\texttt{LL-SUM}) likelihood-based baselines on two qualitatively different multiple-choice tasks.
    \item We show that self-verification is clearly useful on ARC-Challenge for \texttt{Phi-2} and the \texttt{Qwen} models, but much less reliable on TruthfulQA-MC, where some smaller models become prompt-sensitive and several models degrade relative to \texttt{LL-AVG}.
    \item We show that model family and training recipe matter in addition to parameter count for self-verification quality, and that \texttt{LL-SUM} materially narrows the circumstances in which an additional self-verification pass is justified.
\end{itemize}

A public repository containing code, prompts, and experiment artifacts will be made available at \url{https://github.com/phalod-aditya/slm-confidence-signals}.

\section{Related Work}

\subsection{Self-evaluation and same-model self-verification}

A growing line of work asks whether language models can assess the correctness of their own outputs. The closest precedent to our setting is the \(P(\mathrm{True})\) and \(P(\mathrm{IK})\) framework of \textcite{kadavath2022languagemodelsmostlyknow}, who show that language models can sometimes produce useful self-evaluation signals when asked to judge whether their own answers are correct. Related work has also explored whether models can express uncertainty directly in natural language rather than through logits, showing that verbalized uncertainty can be calibrated in some settings \parencite{lin2022teachingmodelsexpressuncertainty}. More recent studies of confidence elicitation and verbalized confidence scores likewise find that reliability depends strongly on prompt design and extraction method rather than following a single robust recipe \parencite{xiong2023canllmsexpress,yang2024verbalizedconf}. More broadly, critique- and oversight-based approaches study how one model output can evaluate another, making self-verification a natural candidate for routing, abstention, and lightweight reliability pipelines \parencite{bai2022constitutionalai}.

Our contribution is not to propose a new introspective signal. Instead, we evaluate an existing signal under a stricter comparative standard: when does same-model self-verification improve selective prediction relative to strong one-pass likelihood baselines, specifically LL-AVG and LL-SUM, across contrasting reasoning- and truthfulness-focused benchmarks?

\subsection{Likelihood-based uncertainty estimation and hallucination detection}

Many standard confidence signals arise directly from a model's predictive distribution. In language modeling and multiple-choice prediction, common approaches derive confidence from token probabilities or sequence likelihoods, often with different normalization choices. These signals are attractive because they are inexpensive to compute and readily available at inference time. At the same time, surveys of uncertainty estimation in NLP emphasize that no single uncertainty measure is uniformly reliable across tasks, datasets, or sources of error \parencite{fisch_uncertainty_nlp}.

This concern is especially relevant for hallucination detection. Recent work shows that uncertainty-based estimators can detect some classes of hallucinations while also highlighting that their usefulness is selective rather than universal \parencite{farquhar2024semanticentropy}. That framing is closely aligned with our setting: any practical claim for self-verification depends on whether it adds value beyond signals already available from the model's own output distribution.

\subsection{Selective prediction, abstention, and deployment-oriented evaluation}

In deployed systems, uncertainty matters because it affects downstream decisions about whether to answer, abstain, defer, or escalate. This is the setting of \emph{selective prediction}, where a system may withhold low-confidence outputs in order to reduce risk on the retained subset. In safety-sensitive applications, a confidence signal is valuable not merely because it correlates with correctness, but because it improves risk--coverage trade-offs. Prior work in high-stakes domains has adopted this framing by studying whether uncertainty estimates improve reliability through selective answering rather than through accuracy alone \parencite{singhal2023largelanguagemodelsencode}.

Our evaluation adopts this deployment-oriented perspective. We therefore go beyond correctness ranking to ask whether same-model self-verification actually supports low-risk selective answering, and whether it does so better than simple likelihood-based alternatives.

\section{Methods}

\subsection{Task setting}

We study confidence estimation in a multiple-choice question answering setting. For each question \(q\), a model selects an answer \(\hat{a}\) from a finite set of candidate options \(\{a_1, \dots, a_K\}\). Our goal is to evaluate whether a confidence estimate can distinguish correct from incorrect predictions and whether it can support \emph{selective prediction}, where the model abstains on low-confidence examples to reduce risk on the retained subset.

We evaluate two benchmarks with qualitatively different failure profiles. The first is \textbf{TruthfulQA-MC} \parencite{lin2021truthfulqa}, loaded from \texttt{EleutherAI/truthful\_qa\_mc} using the validation split and the parquet-converted revision \nolinkurl{refs/convert/parquet}. The second is \textbf{ARC-Challenge} \parencite{clark2018thinkyouhavesolvedquestion}, loaded from \texttt{allenai/ai2\_arc} using the explicit \texttt{ARC-Challenge} configuration and the test split. We enforce strict dataset identity for ARC-Challenge and do not allow fallback to the default configuration, since that setting can mix ARC-Easy and ARC-Challenge examples. Examples whose gold answer cannot be mapped to a valid option index are discarded. In the final reported runs, we did not observe any exclusions of this kind. We fix a shuffled evaluation order using seed \(42\) and save that order to disk so that reruns use the same example sequence.

The model does not generate a free-form answer. Instead, each answer option is scored independently under the model, and prediction is performed by comparing option scores.

\subsection{Models}

We evaluate the following open-weight language models:
\begin{itemize}
    \item \texttt{microsoft/phi-2}
    \item \texttt{Qwen/Qwen2.5-1.5B-Instruct}
    \item \texttt{Qwen/Qwen2.5-7B-Instruct}
    \item \texttt{TinyLlama/TinyLlama-1.1B-Chat-v1.0}
    \item \texttt{deepseek-ai/DeepSeek-R1-Distill-Llama-8B}
\end{itemize}

These models span multiple families, scales, and training recipes, allowing us to study not only capability differences but also whether self-verification behavior transfers across families. The \texttt{deepseek-ai/DeepSeek-R1-Distill-Llama-8B} checkpoint is one of the distilled models released with DeepSeek-R1 \parencite{deepseek2025r1}. For readability, we refer to the two Qwen checkpoints as \texttt{Qwen-1.5B} and \texttt{Qwen-7B}, and to the DeepSeek checkpoint as \texttt{DeepSeek-R1-Distill-8B} in tables and figures. In all experiments, the same model is used both to answer the question and, in the self-verification setting, to judge whether its own predicted answer is correct. All evaluations use score-based inference without sampling. We fix random seeds and the shuffled evaluation order for reproducibility.

\subsection{Likelihood-based prediction and confidence}

For each question, we construct a multiple-choice prompt that lists the question, the candidate answers, and an \texttt{Answer:} field; the exact template is given in Appendix~\ref{app:mc_prompt}. Each candidate option \(a_i\) is scored by its autoregressive log-likelihood conditioned on that prompt. Let \(a_i = (t_1, \dots, t_{L_i})\) denote the tokenized form of option \(i\). We compute the unnormalized score
\begin{equation}
s_{\mathrm{sum}}(a_i \mid q) = \sum_{j=1}^{L_i} \log p(t_j \mid q, t_{<j}),
\end{equation}
and its length-normalized counterpart
\begin{equation}
s_{\mathrm{avg}}(a_i \mid q) = \frac{1}{L_i}\sum_{j=1}^{L_i} \log p(t_j \mid q, t_{<j}).
\end{equation}

Prediction under each scoring rule is obtained by selecting the highest-scoring option. We convert the option scores into a distribution over answer choices using a softmax:
\begin{equation}
p_i = \frac{\exp(s(a_i \mid q))}{\sum_{m=1}^{K} \exp(s(a_m \mid q))},
\end{equation}
and predict
\begin{equation}
\hat{a} = \arg\max_i s(a_i \mid q) = \arg\max_i p_i.
\end{equation}

We study two likelihood-based confidence signals. The first is the probability assigned to the predicted option under the length-normalized score:
\begin{equation}
c_{\mathrm{LL\mbox{-}AVG}} = \max_i p_i^{(\mathrm{avg})}.
\end{equation}
The second is the probability assigned to the predicted option under the unnormalized score:
\begin{equation}
c_{\mathrm{LL\mbox{-}SUM}} = \max_i p_i^{(\mathrm{sum})}.
\end{equation}

We refer to these as \textbf{LL-AVG} and \textbf{LL-SUM}, respectively. We treat LL-AVG as the primary baseline because length normalization is a common way to compare options of different lengths, while retaining LL-SUM as an important comparison baseline given its strong empirical performance.

Unless otherwise stated, all self-verification analyses are conditioned on the \emph{LL-AVG prediction}, i.e., the answer \(\hat{a}\) produced under the length-normalized scoring rule.

\subsection{Same-model self-verification}

Our second confidence estimate is a \emph{same-model self-verification} score. After the model predicts \(\hat{a}\) using LL-AVG, we prompt the \emph{same} model to judge whether that answer is correct.

The default verification prompt asks the model whether its proposed answer is correct using a single-token \texttt{True}/\texttt{False} response; the exact template is given in Appendix~\ref{app:sv_prompt_default}. The prompt ends with a trailing space after \texttt{Answer: } to stabilize tokenization of the next token. We define self-verification confidence as
\begin{equation}
c_{\mathrm{SV}} = P(\mathrm{True} \mid q, \hat{a}),
\end{equation}
where the probability is computed from the next-token logits at the final prompt position rather than from free-text generation.

To make this computation robust across tokenizers, we aggregate probability mass over common single-token variants of \texttt{True} and \texttt{False}, including leading-space and uppercase forms. Let \(\mathcal{T}\) and \(\mathcal{F}\) denote the corresponding token sets, and let \(\ell_v\) be the next-token logit for vocabulary item \(v\). We compute
\begin{align}
\ell_{\mathrm{True}} &= \log \sum_{v \in \mathcal{T}} \exp(\ell_v), \\
\ell_{\mathrm{False}} &= \log \sum_{v \in \mathcal{F}} \exp(\ell_v),
\end{align}
and define
\begin{equation}
c_{\mathrm{SV}} = \sigma\!\left(\ell_{\mathrm{True}} - \ell_{\mathrm{False}}\right),
\end{equation}
where \(\sigma(\cdot)\) is the logistic sigmoid.

If no usable True/False tokenization is available, the implementation falls back to \(\{1,0\}\) token variants, though this fallback serves only as a safeguard rather than the main evaluation path.

\subsection{Prompt ablation for self-verification}

Because self-verification is implemented through an auxiliary prompt, its behavior may depend on prompt wording rather than on the underlying correctness signal alone. We therefore evaluate two verification prompts while keeping the answer-prediction stage fixed. In all prompt-ablation experiments, the model first predicts an answer using LL-AVG, and only the verification prompt is changed.

The default prompt is described above; the audit-style variant reframes the task as answer auditing while preserving the same \texttt{True}/\texttt{False} output format. The exact audit-style template is given in Appendix~\ref{app:sv_prompt_audit}.

For each model--dataset pair, we report self-verification performance separately for each prompt variant and compare them using correctness-ranking and selective-prediction metrics.

\subsection{Evaluation}

We evaluate each confidence signal along three main dimensions.

\paragraph{Prediction accuracy.}
For each example, we record whether the predicted option matches the gold answer under both likelihood-based scoring rules. In the main paper, we report LL-AVG prediction accuracy, while LL-SUM is compared primarily through ranking, calibration, and selective-prediction metrics.

\paragraph{Correctness ranking.}
We measure how well a confidence estimate ranks correct predictions above incorrect predictions using AUROC. AUROC can be interpreted as the probability that a randomly chosen correct prediction receives higher confidence than a randomly chosen incorrect prediction. Because Self-Verify is defined over the answer selected by LL-AVG, AUROC for \(c_{\mathrm{LL\mbox{-}AVG}}\) and \(c_{\mathrm{SV}}\) is computed against the LL-AVG correctness label \(y_{\mathrm{avg}} \in \{0,1\}\). LL-SUM is evaluated on its own predicted answers and corresponding correctness labels \(y_{\mathrm{sum}}\).

\paragraph{Selective prediction.}
To evaluate abstention behavior, we also treat each signal as a selective-prediction score. Examples are sorted by decreasing confidence, and for each retained-prefix coverage level we compute the error rate on the retained subset, yielding a discrete risk--coverage curve. AURC is then computed over all retained-prefix operating points by trapezoidal integration of that curve after prepending the point \((0,1)\); lower values are better. No additional tie-specific correction is applied beyond the induced confidence ordering. From this curve, we report:
\begin{itemize}
    \item area under the risk--coverage curve (AURC; lower is better),
    \item error at \(80\%\) coverage,
    \item error at \(50\%\) coverage,
    \item coverage at \(\leq 20\%\) error,
    \item coverage at \(\leq 10\%\) error.
\end{itemize}

As supporting analyses, we also report Brier score and expected calibration error with 10 bins (ECE-10). For LL-AVG and Self-Verify, these quantities are computed against \(y_{\mathrm{avg}}\); for LL-SUM, they are computed against \(y_{\mathrm{sum}}\).

\subsection{Additional baseline comparisons}

To contextualize the main comparison, we also compute several auxiliary confidence baselines from the option-level multiple-choice distributions. These include the probability margin between the top two answer choices, an entropy-based confidence score defined as one minus normalized predictive entropy, and a temperature-scaled variant of LL-AVG. Temperature scaling is fit separately for each model--dataset pair on a single held-out calibration subset comprising \(20\%\) of examples, with a minimum of 50 calibration examples, sampled once with seed \(42\) and optimized by minimizing negative log-likelihood over the multiple-choice scores. We report these auxiliary baselines only as supporting comparisons. The central comparisons remain LL-AVG and LL-SUM, because both are available from the same answer-scoring pass and therefore define the practical bar that an additional self-verification pass must clear.

\subsection{Statistical testing}

To assess whether self-verification meaningfully changes correctness ranking relative to LL-AVG, we estimate bootstrap confidence intervals for
\begin{equation}
\Delta \mathrm{AUROC}
=
\mathrm{AUROC}(c_{\mathrm{SV}}, y_{\mathrm{avg}})
-
\mathrm{AUROC}(c_{\mathrm{LL\mbox{-}AVG}}, y_{\mathrm{avg}}).
\end{equation}

For each model--dataset pair, we resample evaluation examples with replacement and recompute \(\Delta \mathrm{AUROC}\) over 2000 bootstrap replicates using seed \(42\). Replicates containing only one class are discarded. We report \(95\%\) confidence intervals using the empirical 2.5th and 97.5th percentiles.

\subsection{Implementation details}

All experiments are run in a batched, resumable pipeline. We checkpoint outputs every 100 evaluation examples and use GPU batch sizes of 8 for both multiple-choice likelihood evaluation and self-verification. Maximum sequence length is capped at 256 tokens for both stages.

Models are loaded in \texttt{float16} by default. For larger models, the implementation enables 4-bit NF4 quantization via BitsAndBytes when CUDA is available, with double quantization and \texttt{float16}/\texttt{bfloat16} compute as appropriate. Tokenizer pad tokens are patched to EOS when missing. For \texttt{Phi-2}, we explicitly patch \texttt{pad\_token\_id} in the model configuration before loading to avoid configuration mismatches.

For each run, we save per-example outputs, per-run metrics, the exact dataset configuration, model identifier, seed, prompt variant, and the token-id sets used for True/False self-verification.
\section{Results}

\begin{figure}[t]
    \centering
    \begin{subfigure}[t]{0.48\textwidth}
        \centering
        \includegraphics[width=\linewidth]{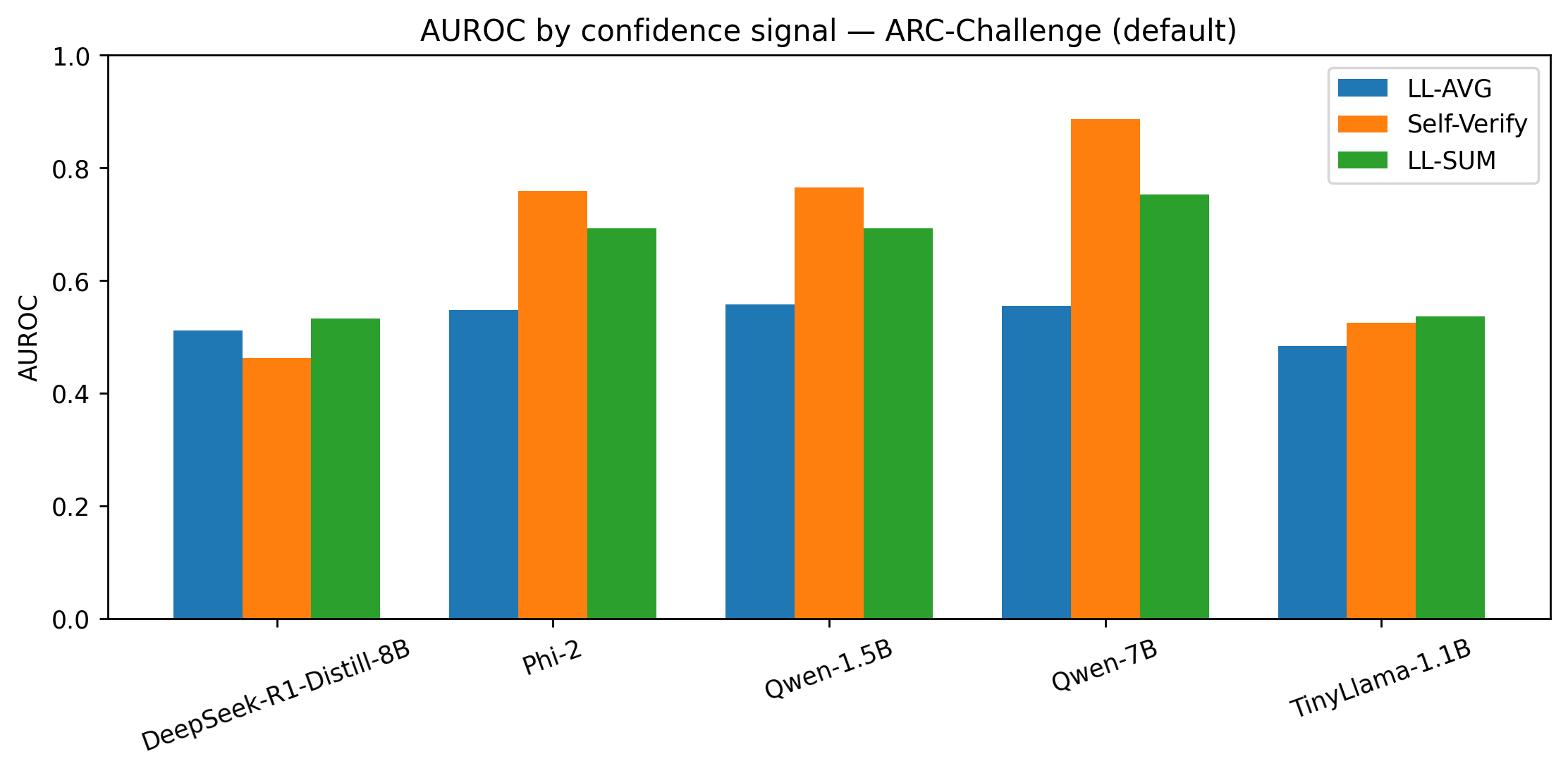}
        \caption{ARC-Challenge (default prompt)}
    \end{subfigure}
    \hfill
    \begin{subfigure}[t]{0.48\textwidth}
        \centering
        \includegraphics[width=\linewidth]{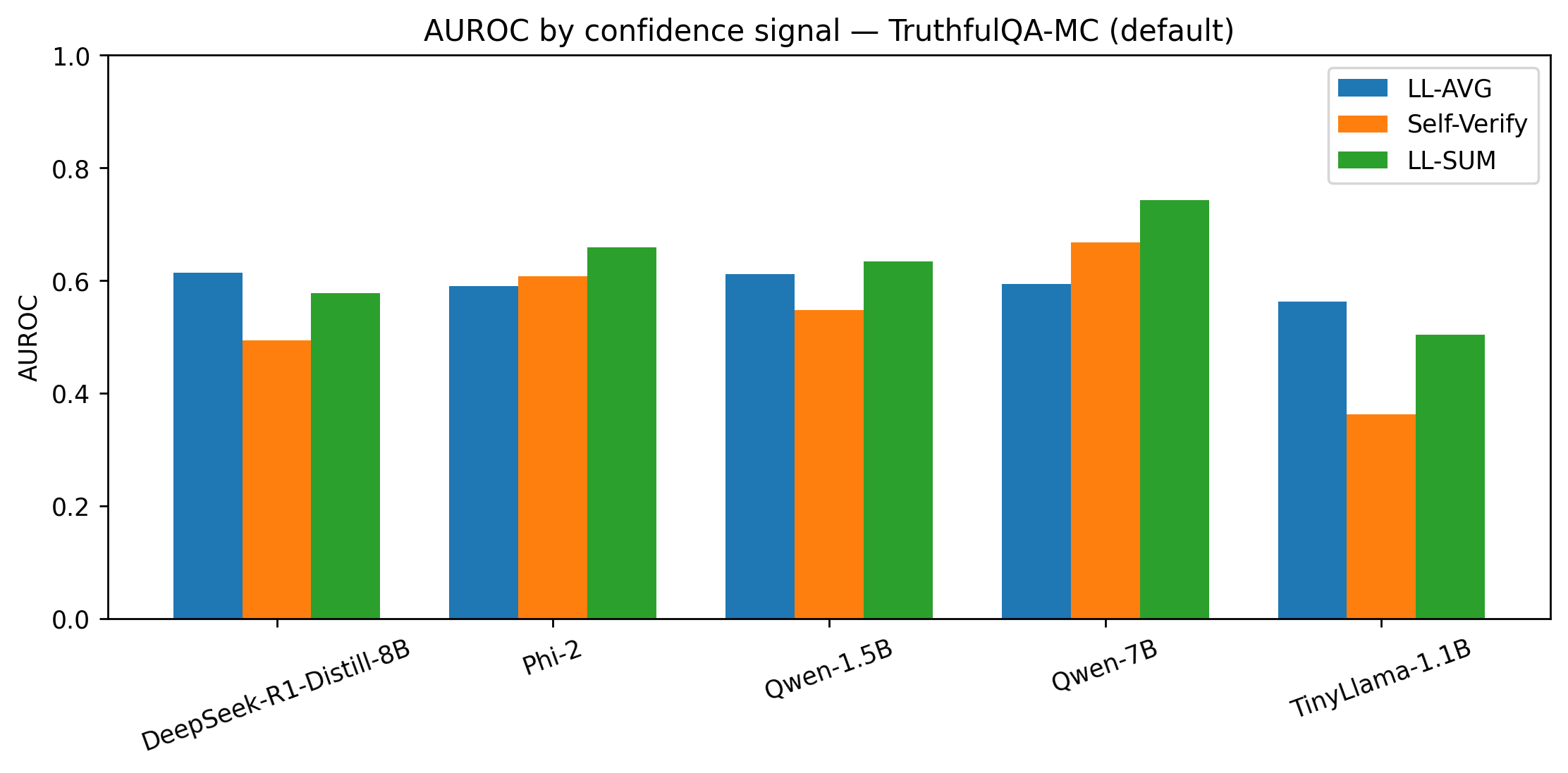}
        \caption{TruthfulQA-MC (default prompt)}
    \end{subfigure}
    \caption{AUROC by confidence signal across datasets under the default verification prompt. Self-Verify is strongly positive on ARC-Challenge for several models, but much less uniform on TruthfulQA-MC.}
    \label{fig:auroc_main}
\end{figure}

\begin{figure}[t]
    \centering
    \begin{subfigure}[t]{0.48\textwidth}
        \centering
        \includegraphics[width=\linewidth]{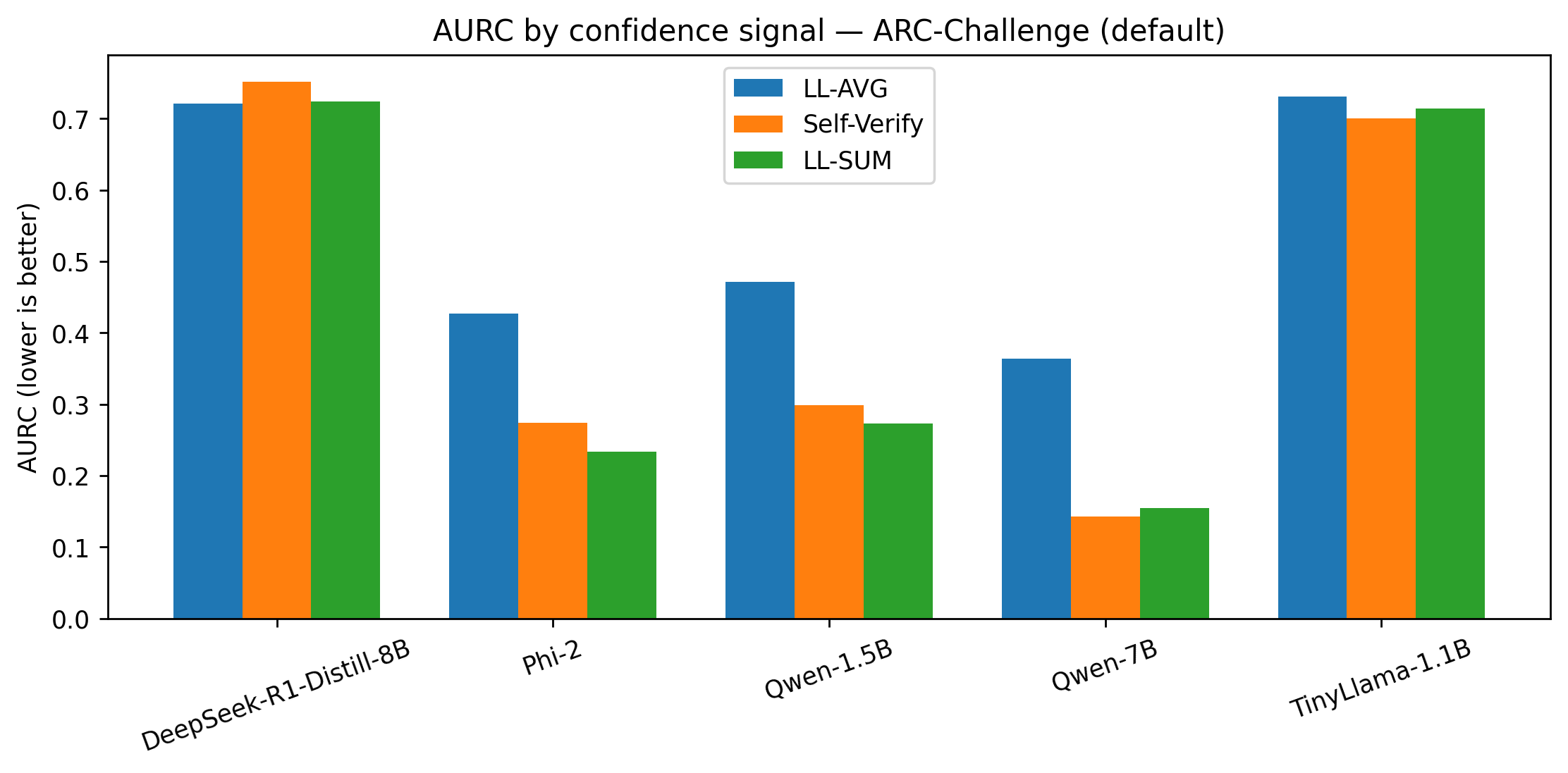}
        \caption{ARC-Challenge (default prompt)}
    \end{subfigure}
    \hfill
    \begin{subfigure}[t]{0.48\textwidth}
        \centering
        \includegraphics[width=\linewidth]{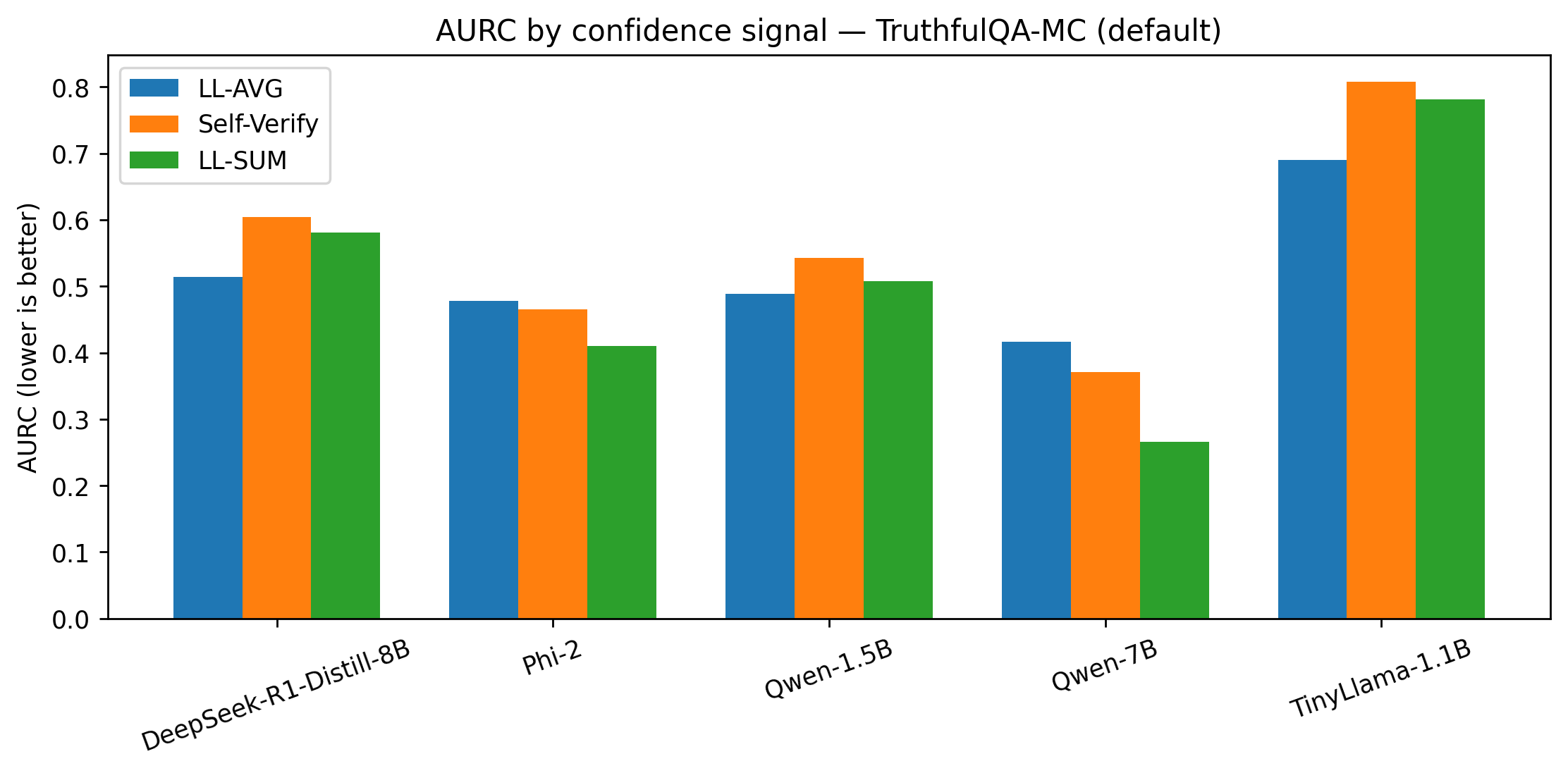}
        \caption{TruthfulQA-MC (default prompt)}
    \end{subfigure}
    \caption{AURC by confidence signal across datasets under the default verification prompt. Lower is better. ARC-Challenge shows large selective-prediction gains for Self-Verify on several models, while TruthfulQA-MC is much less favorable.}
    \label{fig:aurc_main}
\end{figure}

\begin{table*}[t]
    \centering
    \small
    \setlength{\tabcolsep}{4pt}
    \caption{Main results across datasets, models, and prompt variants. The table reports prediction accuracy under LL-AVG, along with AUROC and AURC for LL-AVG, Self-Verify, and LL-SUM. LL-AVG and LL-SUM values are repeated across prompt rows because only the verification prompt is varied. Lower AURC is better.}
    \label{tab:main_results}
    \resizebox{\textwidth}{!}{%
        \begin{tabular}{lllrrrrrrr}
\toprule
Dataset & Model & Prompt & Acc (LL-AVG) & AUROC (LL-AVG) & AUROC (Self-Verify) & AUROC (LL-SUM) & AURC (LL-AVG) & AURC (Self-Verify) & AURC (LL-SUM) \\
\midrule
ARC-Challenge & DeepSeek-R1-Distill-8B & audit\_v1 & 0.278 & 0.511 & 0.489 & 0.533 & 0.722 & 0.738 & 0.724 \\
ARC-Challenge & DeepSeek-R1-Distill-8B & default   & 0.278 & 0.511 & 0.463 & 0.533 & 0.722 & 0.751 & 0.724 \\
ARC-Challenge & Phi-2                  & audit\_v1 & 0.534 & 0.547 & 0.755 & 0.693 & 0.427 & 0.278 & 0.233 \\
ARC-Challenge & Phi-2                  & default   & 0.534 & 0.547 & 0.759 & 0.693 & 0.427 & 0.275 & 0.233 \\
ARC-Challenge & Qwen-1.5B              & audit\_v1 & 0.492 & 0.557 & 0.770 & 0.693 & 0.472 & 0.296 & 0.273 \\
ARC-Challenge & Qwen-1.5B              & default   & 0.492 & 0.557 & 0.765 & 0.693 & 0.472 & 0.298 & 0.273 \\
ARC-Challenge & Qwen-7B                & audit\_v1 & 0.602 & 0.555 & 0.879 & 0.753 & 0.364 & 0.149 & 0.154 \\
ARC-Challenge & Qwen-7B                & default   & 0.602 & 0.555 & 0.886 & 0.753 & 0.364 & 0.143 & 0.154 \\
ARC-Challenge & TinyLlama-1.1B         & audit\_v1 & 0.279 & 0.484 & 0.525 & 0.537 & 0.731 & 0.702 & 0.714 \\
ARC-Challenge & TinyLlama-1.1B         & default   & 0.279 & 0.484 & 0.525 & 0.537 & 0.731 & 0.700 & 0.714 \\
TruthfulQA-MC & DeepSeek-R1-Distill-8B & audit\_v1 & 0.401 & 0.614 & 0.524 & 0.578 & 0.514 & 0.597 & 0.581 \\
TruthfulQA-MC & DeepSeek-R1-Distill-8B & default   & 0.401 & 0.614 & 0.494 & 0.578 & 0.514 & 0.605 & 0.581 \\
TruthfulQA-MC & Phi-2                  & audit\_v1 & 0.468 & 0.590 & 0.624 & 0.659 & 0.478 & 0.461 & 0.410 \\
TruthfulQA-MC & Phi-2                  & default   & 0.468 & 0.590 & 0.608 & 0.659 & 0.478 & 0.465 & 0.410 \\
TruthfulQA-MC & Qwen-1.5B              & audit\_v1 & 0.439 & 0.611 & 0.620 & 0.634 & 0.488 & 0.500 & 0.508 \\
TruthfulQA-MC & Qwen-1.5B              & default   & 0.439 & 0.611 & 0.548 & 0.634 & 0.488 & 0.543 & 0.508 \\
TruthfulQA-MC & Qwen-7B                & audit\_v1 & 0.529 & 0.593 & 0.665 & 0.742 & 0.416 & 0.367 & 0.266 \\
TruthfulQA-MC & Qwen-7B                & default   & 0.529 & 0.593 & 0.667 & 0.742 & 0.416 & 0.371 & 0.266 \\
TruthfulQA-MC & TinyLlama-1.1B         & audit\_v1 & 0.265 & 0.562 & 0.372 & 0.503 & 0.691 & 0.800 & 0.782 \\
TruthfulQA-MC & TinyLlama-1.1B         & default   & 0.265 & 0.562 & 0.363 & 0.503 & 0.691 & 0.807 & 0.782 \\
\bottomrule
\end{tabular}
    }
\end{table*}

\subsection{Overall comparison across datasets and models}

Table~\ref{tab:main_results} and Figures~\ref{fig:auroc_main}--\ref{fig:aurc_main} show a sharp contrast between ARC-Challenge and TruthfulQA-MC. Additional AUROC and AURC breakdowns across datasets and prompt variants are shown in Appendix~\ref{app:figures}. On ARC-Challenge, same-model self-verification often improves substantially over LL-AVG. The strongest result appears in \texttt{Qwen-7B}, where Self-Verify improves AUROC from \(0.555\) to \(0.886\) under the default prompt and reduces AURC from \(0.364\) to \(0.143\). \texttt{Phi-2} and \texttt{Qwen-1.5B} show the same qualitative pattern, while \texttt{TinyLlama-1.1B} is only weakly positive and \texttt{DeepSeek-R1-Distill-8B} is negative under both prompts.

TruthfulQA-MC is much less favorable. The clearest positive case is again \texttt{Qwen-7B}, which improves from \(0.593\) to \(0.667\) in AUROC and reduces AURC from \(0.416\) to \(0.371\). \texttt{Phi-2} is mildly positive. \texttt{Qwen-1.5B} is prompt-sensitive: under the audit-style prompt it is slightly better than LL-AVG in AUROC (\(0.620\) vs.\ \(0.611\)), but under the default prompt it drops to \(0.548\). The negative cases are substantial: \texttt{TinyLlama-1.1B} falls from \(0.562\) to \(0.363\) in AUROC under the default prompt, and \texttt{DeepSeek-R1-Distill-8B} falls from \(0.614\) to \(0.494\). The main table therefore already shows the paper's central tension: ARC-Challenge contains clear wins for Self-Verify over LL-AVG, but TruthfulQA-MC and the presence of strong LL-SUM results prevent a broad practical claim.

\subsection{Self-Verify relative to LL-AVG}

Our primary comparison is between Self-Verify and LL-AVG, since both are defined relative to the same LL-AVG answer prediction. On ARC-Challenge, Self-Verify clearly outperforms LL-AVG for \texttt{Qwen-1.5B}, \texttt{Qwen-7B}, and \texttt{Phi-2}. The AUROC gains are large: about \(+0.21\) for \texttt{Qwen-1.5B}, \(+0.21\) for \texttt{Phi-2}, and \(+0.33\) for \texttt{Qwen-7B}, depending on prompt. These are matched by substantial AURC reductions of about \(-0.17\), \(-0.15\), and \(-0.22\), respectively. \texttt{TinyLlama-1.1B} improves only marginally, while \texttt{DeepSeek-R1-Distill-8B} is negative under both prompts.

TruthfulQA-MC produces a much more heterogeneous picture. \texttt{Qwen-7B} remains consistently positive relative to LL-AVG, improving AUROC by about \(+0.07\) under both prompts and reducing AURC by about \(0.05\). \texttt{Phi-2} is only mildly positive, with AUROC gains of \(+0.034\) and \(+0.018\). \texttt{Qwen-1.5B}, however, is prompt-sensitive: the audit-style prompt yields a near-neutral improvement in AUROC (\(+0.009\)), whereas the default prompt becomes clearly negative (\(-0.063\)). The negative cases are stronger still. On TruthfulQA-MC, \texttt{TinyLlama-1.1B} drops by about \(0.19\) to \(0.20\) AUROC relative to LL-AVG, and \texttt{DeepSeek-R1-Distill-8B} drops by about \(0.09\) to \(0.12\).

Relative to LL-AVG, Self-Verify is clearly useful in some ARC-Challenge settings. In the more heterogeneous TruthfulQA-MC regime, its value becomes conditional on model family and prompt wording.

\begin{table*}[t]
    \centering
    \small
    \setlength{\tabcolsep}{4pt}
    \caption{Pairwise deltas for Self-Verify relative to LL-AVG and LL-SUM. Negative \(\Delta\)AURC indicates lower risk for Self-Verify.}
    \label{tab:deltas}
    \resizebox{\textwidth}{!}{%
        \begin{tabular}{lllrrrr}
\toprule
Dataset & Model & Prompt & $\Delta$AUROC (SV--LLAVG) & $\Delta$AURC (SV--LLAVG) & $\Delta$AUROC (SV--LLSUM) & $\Delta$AURC (SV--LLSUM) \\
\midrule
ARC-Challenge & DeepSeek-R1-Distill-8B & audit\_v1 & -0.022 &  0.016 & -0.043 &  0.014 \\
ARC-Challenge & DeepSeek-R1-Distill-8B & default   & -0.048 &  0.030 & -0.069 &  0.027 \\
ARC-Challenge & Phi-2                  & audit\_v1 &  0.207 & -0.149 &  0.061 &  0.045 \\
ARC-Challenge & Phi-2                  & default   &  0.211 & -0.153 &  0.066 &  0.041 \\
ARC-Challenge & Qwen-1.5B              & audit\_v1 &  0.213 & -0.176 &  0.077 &  0.023 \\
ARC-Challenge & Qwen-1.5B              & default   &  0.207 & -0.173 &  0.071 &  0.025 \\
ARC-Challenge & Qwen-7B                & audit\_v1 &  0.324 & -0.215 &  0.126 & -0.006 \\
ARC-Challenge & Qwen-7B                & default   &  0.331 & -0.221 &  0.134 & -0.011 \\
ARC-Challenge & TinyLlama-1.1B         & audit\_v1 &  0.041 & -0.029 & -0.012 & -0.012 \\
ARC-Challenge & TinyLlama-1.1B         & default   &  0.041 & -0.031 & -0.012 & -0.014 \\
TruthfulQA-MC & DeepSeek-R1-Distill-8B & audit\_v1 & -0.090 &  0.084 & -0.053 &  0.016 \\
TruthfulQA-MC & DeepSeek-R1-Distill-8B & default   & -0.120 &  0.091 & -0.084 &  0.023 \\
TruthfulQA-MC & Phi-2                  & audit\_v1 &  0.034 & -0.018 & -0.035 &  0.051 \\
TruthfulQA-MC & Phi-2                  & default   &  0.018 & -0.013 & -0.051 &  0.055 \\
TruthfulQA-MC & Qwen-1.5B              & audit\_v1 &  0.009 &  0.012 & -0.013 & -0.008 \\
TruthfulQA-MC & Qwen-1.5B              & default   & -0.063 &  0.055 & -0.086 &  0.035 \\
TruthfulQA-MC & Qwen-7B                & audit\_v1 &  0.072 & -0.049 & -0.078 &  0.101 \\
TruthfulQA-MC & Qwen-7B                & default   &  0.074 & -0.045 & -0.075 &  0.104 \\
TruthfulQA-MC & TinyLlama-1.1B         & audit\_v1 & -0.190 &  0.109 & -0.131 &  0.018 \\
TruthfulQA-MC & TinyLlama-1.1B         & default   & -0.199 &  0.117 & -0.141 &  0.026 \\
\bottomrule
\end{tabular}
    }
\end{table*}

\begin{table*}[t]
    \centering
    \small
    \setlength{\tabcolsep}{4pt}
    \caption{Bootstrap estimates for \(\Delta\)AUROC \(=\mathrm{AUROC}(\mathrm{Self\mbox{-}Verify})-\mathrm{AUROC}(\mathrm{LL\mbox{-}AVG})\). Positive values indicate that Self-Verify ranks correct predictions above incorrect predictions better than LL-AVG. Intervals are not corrected for multiple comparisons and should be interpreted as exploratory.}
    \label{tab:delta_auroc}
    \resizebox{\textwidth}{!}{%
        \begin{tabular}{lllrrr}
\toprule
Dataset & Model & Prompt & Mean $\Delta$AUROC & 2.5\% CI & 97.5\% CI \\
\midrule
ARC-Challenge & DeepSeek-R1-Distill-8B & audit\_v1 & -0.023 & -0.073 &  0.029 \\
ARC-Challenge & DeepSeek-R1-Distill-8B & default   & -0.049 & -0.100 &  0.002 \\
ARC-Challenge & Phi-2                  & audit\_v1 &  0.208 &  0.168 &  0.251 \\
ARC-Challenge & Phi-2                  & default   &  0.212 &  0.173 &  0.253 \\
ARC-Challenge & Qwen-1.5B              & audit\_v1 &  0.213 &  0.169 &  0.257 \\
ARC-Challenge & Qwen-1.5B              & default   &  0.208 &  0.164 &  0.251 \\
ARC-Challenge & Qwen-7B                & audit\_v1 &  0.323 &  0.284 &  0.362 \\
ARC-Challenge & Qwen-7B                & default   &  0.331 &  0.293 &  0.369 \\
ARC-Challenge & TinyLlama-1.1B         & audit\_v1 &  0.042 & -0.013 &  0.092 \\
ARC-Challenge & TinyLlama-1.1B         & default   &  0.041 & -0.015 &  0.095 \\
TruthfulQA-MC & DeepSeek-R1-Distill-8B & audit\_v1 & -0.091 & -0.154 & -0.031 \\
TruthfulQA-MC & DeepSeek-R1-Distill-8B & default   & -0.121 & -0.183 & -0.060 \\
TruthfulQA-MC & Phi-2                  & audit\_v1 &  0.034 & -0.022 &  0.086 \\
TruthfulQA-MC & Phi-2                  & default   &  0.018 & -0.038 &  0.071 \\
TruthfulQA-MC & Qwen-1.5B              & audit\_v1 &  0.010 & -0.046 &  0.063 \\
TruthfulQA-MC & Qwen-1.5B              & default   & -0.063 & -0.120 & -0.008 \\
TruthfulQA-MC & Qwen-7B                & audit\_v1 &  0.071 &  0.017 &  0.128 \\
TruthfulQA-MC & Qwen-7B                & default   &  0.073 &  0.020 &  0.129 \\
TruthfulQA-MC & TinyLlama-1.1B         & audit\_v1 & -0.190 & -0.264 & -0.116 \\
TruthfulQA-MC & TinyLlama-1.1B         & default   & -0.199 & -0.273 & -0.128 \\
\bottomrule
\end{tabular}
    }
\end{table*}

Table~\ref{tab:deltas} makes the asymmetry across datasets more explicit. The largest positive Self-Verify deltas relative to LL-AVG occur on ARC-Challenge for \texttt{Phi-2}, \texttt{Qwen-1.5B}, and especially \texttt{Qwen-7B}. The same table also shows that these gains do not generalize uniformly across datasets or baselines: on TruthfulQA-MC, the deltas are much smaller, more prompt-sensitive, and often negative; and even on ARC-Challenge, LL-SUM remains competitive in AURC for some models. This prevents an overly broad reading of the ARC gains.

The bootstrap results in Table~\ref{tab:delta_auroc} reinforce this picture. On ARC-Challenge, the gains for \texttt{Phi-2}, \texttt{Qwen-1.5B}, and \texttt{Qwen-7B} are clearly separated from zero, with 95\% bootstrap intervals entirely positive. The strongest improvement is \texttt{Qwen-7B}, with \(\Delta\)AUROC \(=0.331\) under the default prompt. By contrast, \texttt{TinyLlama-1.1B} on ARC-Challenge has only a small gain and its interval overlaps zero, while \texttt{DeepSeek-R1-Distill-8B} remains negative or near zero. On TruthfulQA-MC, \texttt{Qwen-7B} is the only clearly positive case, with 95\% bootstrap intervals fully above zero under both prompt variants. \texttt{Phi-2} is weak, \texttt{Qwen-1.5B} is mixed and prompt-sensitive, and \texttt{TinyLlama-1.1B} and \texttt{DeepSeek-R1-Distill-8B} are robustly negative. Thus, the strongest improvements and the strongest failures are both large enough to be practically meaningful rather than minor fluctuations around the LL-AVG baseline.

\subsection{Selective prediction and risk--coverage behavior}

\begin{figure}[t]
    \centering
    \includegraphics[width=\linewidth]{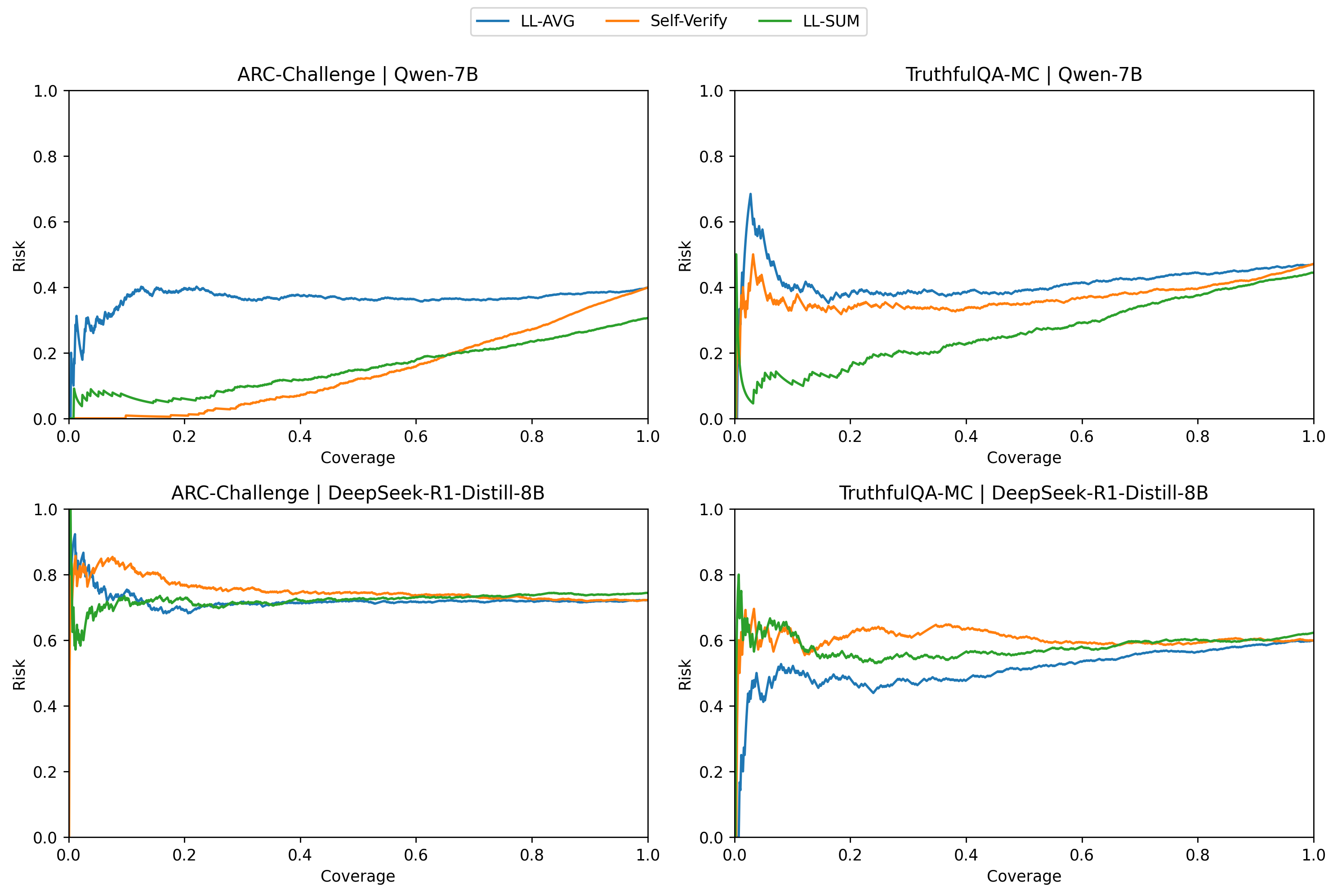}
    \caption{Representative risk--coverage curves. ARC-Challenge shows clear selective-prediction gains for Self-Verify relative to LL-AVG, while TruthfulQA-MC is much less stable and more prompt-sensitive.}
    \label{fig:risk_coverage_examples}
\end{figure}

Because our motivating use case is abstention, the key deployment metric is not AUROC alone but selective prediction behavior. Table~\ref{tab:operating_points} reports representative operating points for the default prompt, Figure~\ref{fig:risk_coverage_examples} shows representative risk--coverage curves, and Appendix Table~\ref{tab:operating_points_appendix} reports the full set of operating-point results. On ARC-Challenge, self-verification substantially improves risk--coverage trade-offs relative to LL-AVG for the strongest positive models. For \texttt{Qwen-7B}, Self-Verify reduces error at \(80\%\) coverage from \(0.369\) to \(0.271\) and raises coverage at \(10\%\) error from \(0.009\) to \(0.463\). For \texttt{Qwen-1.5B}, error at \(50\%\) coverage drops from \(0.462\) to \(0.321\), and coverage at \(10\%\) error rises from \(0.002\) to \(0.096\). \texttt{Phi-2} shows the same qualitative pattern, though the gains are smaller than for \texttt{Qwen-7B}.

\begin{table}[t]
    \centering
    \small
    \setlength{\tabcolsep}{4pt}
    \caption{Representative selective-prediction operating points under the default prompt for the models emphasized in the main text. Lower error at fixed coverage and higher coverage at fixed error indicate better abstention behavior. Full results appear in Appendix Table~\ref{tab:operating_points_appendix}.}
    \label{tab:operating_points}
    \resizebox{0.96\linewidth}{!}{%
        \begin{tabular}{lllrrrr}
\toprule
Dataset & Model & Signal & err@80\%cov & err@50\%cov & cov@20\%err & cov@10\%err \\
\midrule
ARC-Challenge & Phi-2     & LL-AVG      & 0.447 & 0.451 & 0.004 & 0.002 \\
ARC-Challenge & Phi-2     & Self-Verify & 0.387 & 0.268 & 0.317 & 0.073 \\
ARC-Challenge & Phi-2     & LL-SUM      & 0.327 & 0.253 & 0.412 & 0.061 \\
ARC-Challenge & Qwen-1.5B & LL-AVG      & 0.485 & 0.462 & 0.007 & 0.002 \\
ARC-Challenge & Qwen-1.5B & Self-Verify & 0.431 & 0.321 & 0.262 & 0.096 \\
ARC-Challenge & Qwen-1.5B & LL-SUM      & 0.374 & 0.299 & 0.241 & 0.094 \\
ARC-Challenge & Qwen-7B   & LL-AVG      & 0.369 & 0.362 & 0.026 & 0.009 \\
ARC-Challenge & Qwen-7B   & Self-Verify & 0.271 & 0.121 & 0.666 & 0.463 \\
ARC-Challenge & Qwen-7B   & LL-SUM      & 0.235 & 0.148 & 0.679 & 0.334 \\
TruthfulQA-MC & Phi-2     & LL-AVG      & 0.501 & 0.456 & 0.009 & 0.006 \\
TruthfulQA-MC & Phi-2     & Self-Verify & 0.494 & 0.430 & 0.012 & 0.007 \\
TruthfulQA-MC & Phi-2     & LL-SUM      & 0.521 & 0.442 & 0.126 & 0.019 \\
TruthfulQA-MC & Qwen-1.5B & LL-AVG      & 0.528 & 0.491 & 0.000 & 0.000 \\
TruthfulQA-MC & Qwen-1.5B & Self-Verify & 0.545 & 0.523 & 0.000 & 0.000 \\
TruthfulQA-MC & Qwen-1.5B & LL-SUM      & 0.589 & 0.538 & 0.010 & 0.003 \\
TruthfulQA-MC & Qwen-7B   & LL-AVG      & 0.444 & 0.392 & 0.007 & 0.004 \\
TruthfulQA-MC & Qwen-7B   & Self-Verify & 0.395 & 0.351 & 0.009 & 0.003 \\
TruthfulQA-MC & Qwen-7B   & LL-SUM      & 0.375 & 0.257 & 0.351 & 0.118 \\
\bottomrule
\end{tabular}

    }
\end{table}

On TruthfulQA-MC, the selective-prediction story is weaker. \texttt{Qwen-7B} still improves over LL-AVG, reducing error at \(80\%\) coverage from \(0.444\) to \(0.395\) under the default prompt, but the gain is modest and still leaves Self-Verify well behind LL-SUM, which reaches \(0.375\) at \(80\%\) coverage and \(0.257\) at \(50\%\) coverage while also achieving much higher coverage at low error. \texttt{Phi-2} is again only mildly positive. \texttt{Qwen-1.5B} is unstable across prompts and never achieves meaningful low-error coverage. \texttt{TinyLlama-1.1B} and \texttt{DeepSeek-R1-Distill-8B} both become worse under Self-Verify. This matters directly for deployment: a second-stage self-check can still be harmful if it assigns relatively higher confidence to wrong answers in a truthfulness-intensive regime.

The risk--coverage evidence sharpens the main empirical story. On ARC-Challenge, Self-Verify is not merely a ranking signal that looks better on AUROC; it also yields meaningful abstention gains relative to LL-AVG. But these gains do not transfer uniformly to TruthfulQA-MC, where the signal is less dependable and often remains substantially weaker than LL-SUM.

\subsection{Prompt sensitivity and family effects}

Prompt sensitivity and family effects provide a more qualified view of where Self-Verify helps. The prompt-ablation values and supporting audit-style AUROC/AURC comparisons are reported in Appendix Table~\ref{tab:prompt_ablation_appendix} and Appendix Figures~\ref{fig:app_prompt_ablation}--\ref{fig:app_scale_family}.

On ARC-Challenge, Self-Verify is largely robust across the two prompt variants we test. The AUROC difference between the prompts is at most \(0.026\), and for the strongest positive models it is smaller: \(0.004\) for \texttt{Phi-2}, \(0.005\) for \texttt{Qwen-1.5B}, and \(0.008\) for \texttt{Qwen-7B}. The corresponding AURC shifts are also small. TruthfulQA-MC is more sensitive. The clearest example is \texttt{Qwen-1.5B}, where Self-Verify AUROC shifts from \(0.620\) under the audit-style prompt to \(0.548\) under the default prompt, while AURC worsens from \(0.500\) to \(0.543\). \texttt{DeepSeek-R1-Distill-8B} also shifts under prompt variation, though it remains negative overall, with AUROC moving from \(0.524\) to \(0.494\). By contrast, \texttt{Qwen-7B} remains relatively stable even on TruthfulQA-MC, with AUROC near \(0.665\) to \(0.667\) across prompts.

Within the Qwen family, moving from \texttt{Qwen-1.5B} to \texttt{Qwen-7B} leaves LL-AVG AUROC on ARC-Challenge essentially unchanged (\(0.557\) vs.\ \(0.555\)), yet Self-Verify improves from about \(0.77\) to about \(0.89\). The same pattern appears in AURC: LL-AVG improves from \(0.472\) to \(0.364\), but Self-Verify improves more sharply, from about \(0.30\) to about \(0.14\). On TruthfulQA-MC, the same within-family pattern appears more weakly: \texttt{Qwen-1.5B} is mixed and prompt-sensitive, while \texttt{Qwen-7B} remains positive relative to LL-AVG under both prompts. \texttt{DeepSeek-R1-Distill-8B} makes the cross-family comparison more informative. Despite similar scale to \texttt{Qwen-7B}, it does not behave like a strong self-verifier in this setup. On ARC-Challenge it underperforms LL-AVG, and on TruthfulQA-MC it degrades substantially. The contrast does not establish a mechanism, but it does argue against a simple monotonic scaling story. Within the tested model set, prompt robustness and self-verification quality improve clearly within Qwen, but do not transfer uniformly across families.

\subsection{LL-SUM as a competing baseline}

From a deployment perspective, the comparison to LL-SUM is at least as important as the comparison to LL-AVG, because LL-SUM requires no second verification pass. Table~\ref{tab:deltas} shows that on ARC-Challenge, Self-Verify usually improves over LL-SUM in AUROC. For example, \texttt{Qwen-7B} improves from \(0.753\) under LL-SUM to \(0.886\) under Self-Verify, and \texttt{Phi-2} improves from \(0.693\) to \(0.759\). In AURC, however, the comparison is more nuanced. For \texttt{Qwen-7B}, Self-Verify is slightly better than LL-SUM (\(0.143\) vs.\ \(0.154\)); for \texttt{Qwen-1.5B}, LL-SUM remains better in AURC (\(0.273\) vs.\ \(0.298\)); and for \texttt{Phi-2}, LL-SUM is clearly better in AURC (\(0.233\) vs.\ \(0.275\)) despite Self-Verify having the higher AUROC.

The LL-SUM comparison is even more consequential on TruthfulQA-MC. Although Self-Verify improves over LL-AVG for \texttt{Qwen-7B}, it remains well below LL-SUM, which reaches \(0.742\) AUROC and \(0.266\) AURC. The same pattern holds for \texttt{Phi-2}, where LL-SUM outperforms Self-Verify on both AUROC and AURC. For \texttt{Qwen-1.5B}, the audit-style prompt makes Self-Verify slightly better than LL-AVG in AUROC, but it still does not surpass LL-SUM. These comparisons narrow the practical claim: Self-Verify can be useful relative to LL-AVG, especially on ARC-Challenge, but its advantage is harder to defend once a strong one-pass baseline is available.

Appendix Table~\ref{tab:appendix_aux_baselines} adds a useful check. Margin and temperature-scaled LL-AVG are competitive in a few TruthfulQA-MC settings, but they do not overturn the main picture: none consistently matches Self-Verify on ARC-Challenge or LL-SUM on TruthfulQA-MC. Overall, LL-SUM provides an important ceiling in the present setting. The strongest empirical case for Self-Verify is therefore not that it dominates all alternatives, but that it becomes a strong and deployment-relevant signal in some regimes, most clearly ARC-Challenge with stronger Qwen models, while remaining substantially weaker in others.

\FloatBarrier
\section{Discussion}

The main empirical lesson is comparative: self-verification can add value over LL-AVG in some settings, but that value narrows considerably once stronger one-pass baselines such as LL-SUM are taken seriously.

The clearest positive regime is ARC-Challenge. For several models, especially \texttt{Qwen-7B}, self-verification substantially improves both correctness ranking and selective prediction relative to LL-AVG. These gains are not limited to AUROC; they carry through to lower AURC and better abstention behavior. In this setting, a second-pass self-check can provide useful information about whether an answer should be trusted.

TruthfulQA-MC is less favorable. Here, self-verification is unstable across models and, for some models, across prompt variants. The contrast with ARC-Challenge is consistent with the view that not all errors are equally detectable by same-model self-audit: a model may recognize uncertainty in a fragile reasoning process while still failing to detect that a fluent answer is false. We emphasize, however, that this is an interpretation rather than a demonstrated mechanism. ARC-Challenge and TruthfulQA-MC differ along several dimensions at once, so the contrast should be read as an informative stress test rather than a clean causal decomposition. The prompt sensitivity observed on TruthfulQA-MC is also consistent with recent work showing that elicited confidence depends materially on prompt format \parencite{xiong2023canllmsexpress,yang2024verbalizedconf}.

Within the Qwen family at the tested scales, self-verification improves clearly from \texttt{Qwen-1.5B} to \texttt{Qwen-7B}, especially on ARC-Challenge. However, this pattern does not generalize across families: \texttt{DeepSeek-R1-Distill-8B} yields a weak or harmful self-verification signal relative to LL-AVG on ARC-Challenge and degrades robustly on TruthfulQA-MC despite its similar scale. Scale alone is therefore insufficient to explain self-verification reliability within the tested set. The comparison to LL-SUM narrows the practical claim further. Relative to LL-AVG, self-verification can be highly useful in some regimes. Relative to LL-SUM, the story is less favorable, especially on TruthfulQA-MC. In several settings, LL-SUM remains the stronger confidence signal while requiring only a single scoring pass.

Appendix calibration results add an additional nuance. In some settings, Self-Verify improves AUROC or AURC while worsening Brier score or ECE relative to LL-AVG. Better correctness ranking therefore does not imply better probability calibration. In practice, self-verification may still be useful for abstention while remaining a poor calibrated probability estimate.

Overall, these findings support a cautious view of introspective confidence in language models. Self-verification should not be adopted as a blanket confidence wrapper without task-specific validation.

\section{Limitations}

Our findings should be interpreted in light of several limitations, many of which are consistent with broader observations in the uncertainty and self-evaluation literature.

\subsection{Limited model and family coverage}

Although we evaluate multiple open-weight models spanning several families, our model set remains limited. In particular, we include only a small number of scale points within each family and only a small number of families overall. Prior work on self-evaluation and \emph{P(True)} suggests that confidence quality can vary materially with model scale and evaluation format rather than following a uniform pattern across models \parencite{kadavath2022languagemodelsmostlyknow}. Our results are sufficient to show that self-verification does not follow a simple monotonic relationship with parameter count in the tested set, but they are not sufficient to support a general family-level or scale-law claim. For example, the improvement from \texttt{Qwen-1.5B} to \texttt{Qwen-7B} provides evidence of scale-linked improvement within one family, while the behavior of \texttt{DeepSeek-R1-Distill-8B} shows that similar size does not guarantee similar reliability across families. These observations are therefore descriptive rather than exhaustive. A broader study would require denser coverage within families, more families, and additional models trained with different alignment and distillation recipes.

\subsection{Restriction to multiple-choice evaluation}

Our experiments are conducted entirely in a controlled multiple-choice setting. This makes it possible to define correctness precisely, compare confidence signals cleanly, and evaluate abstention behavior with minimal ambiguity. However, it also limits the scope of our conclusions. Prior work has found that structured true/false or multiple-choice formats can elicit cleaner self-evaluation signals than open-ended generation, while selective generation in free-form settings introduces different failure modes and evaluation challenges \parencite{kadavath2022languagemodelsmostlyknow,ren2023selfevaluation}. Confidence estimation in free-form generation can differ substantially from confidence estimation over a small set of discrete answer options, especially when correctness is partial, underspecified, or difficult to score automatically. In addition, our two-benchmark contrast should not be read as a clean causal decomposition of error type: ARC-Challenge and TruthfulQA-MC differ along several dimensions at once beyond the informal reasoning-vs.-truthfulness framing used in the paper. As a result, our findings should not be taken as evidence that same-model self-verification will behave identically in open-ended generation, tool use, or long-form reasoning tasks, nor that benchmark-level differences isolate a single underlying failure mode.

\subsection{Baseline coverage}

We compare self-verification primarily against two likelihood-based baselines, \textbf{LL-AVG} and \textbf{LL-SUM}. These are strong and informative baselines for our setting, and the comparison to \textbf{LL-SUM} is especially important because it shows that self-verification is not universally preferable even when it improves over \textbf{LL-AVG}. However, our baseline set is not exhaustive. In particular, we do not evaluate other important families of uncertainty estimators, such as semantic-entropy-style methods, ensemble-based uncertainty, self-consistency-based signals, or learned post-hoc confidence models. Recent work on hallucination detection shows that useful uncertainty estimates can depend on moving beyond token-likelihood surrogates alone, for example by reasoning at the level of semantic meaning rather than surface form \parencite{farquhar2024semanticentropy,kossen2024semanticentropyprobes}. Future work should test whether the regime dependence we identify persists relative to these broader alternatives.

\subsection{Limited prompt-ablation scope}

Our prompt-ablation analysis is intended to test whether self-verification depends strongly on prompt wording, but it covers only a small number of prompt variants. This is sufficient to show that prompt sensitivity exists in some settings, especially on TruthfulQA-MC, but it does not provide a comprehensive characterization of prompt effects beyond the two variants we test. Prior work on confidence elicitation likewise finds that verbalized or elicited confidence can vary materially with prompt design, extraction method, and aggregation strategy rather than following a single robust prompting recipe \parencite{xiong2023canllmsexpress,yang2024verbalizedconf}. Different instruction styles, output formats, answer framing choices, or more elaborate verification prompts may produce different results. Accordingly, our conclusions about prompt robustness should be interpreted as evidence of sensitivity rather than as a full map of the prompt design space.

\subsection{Descriptive rather than mechanistic conclusions}

This paper is primarily an empirical characterization study. It identifies where same-model self-verification appears reliable, where it becomes unstable, and where it loses to simple likelihood-based alternatives. It does not, however, establish the mechanism underlying these patterns. This limitation is not unique to our study: recent work on uncertainty estimation and hallucination detection has shown that useful empirical signals can be identified before their causal or representational basis is well understood \parencite{farquhar2024semanticentropy,kossen2024semanticentropyprobes}. In particular, while the contrast between \texttt{Qwen-7B} and \texttt{DeepSeek-R1-Distill-8B} suggests that model family or training recipe matters in addition to scale, our experiments do not isolate whether the relevant factor is distillation, instruction tuning, alignment strategy, pretraining distribution, or some other property of the model. Similarly, while the ARC-Challenge / TruthfulQA-MC contrast is consistent with the idea that different error regimes place different demands on self-verification, our experiments do not prove a specific cognitive or representational explanation for this difference.

Taken together, these limitations suggest that our results should be read as a boundary-mapping study rather than a universal theory of self-verification. The main contribution of the paper is not to claim that same-model self-verification is a generally reliable confidence wrapper, but to show that its usefulness is conditional: task type, model family, prompt formulation, and baseline choice all materially affect whether it helps.
\section{Conclusion}

We evaluated same-model self-verification as a confidence signal for selective prediction against two likelihood-based baselines, LL-AVG and LL-SUM, across ARC-Challenge and TruthfulQA-MC, multiple model families, scales, and prompt variants. The resulting picture is useful but narrow. On ARC-Challenge, self-verification substantially improves correctness ranking and abstention behavior for \texttt{Phi-2} and the \texttt{Qwen} models relative to LL-AVG, with the clearest gains appearing in \texttt{Qwen-7B}. On TruthfulQA-MC, however, it is much less reliable: smaller models can become prompt-sensitive, \texttt{DeepSeek-R1-Distill-8B} is consistently weak in this setup, and \texttt{LL-SUM} remains a strong practical competitor.

The main takeaway is therefore comparative rather than universal. Same-model self-verification can be useful in some regimes, but its value depends on the task setting, model family, prompt formulation, and the baseline it must beat. Future work should test whether this pattern persists across broader model families, stronger uncertainty baselines, and open-ended generation settings beyond multiple-choice evaluation.

\FloatBarrier
\clearpage
\printbibliography
\clearpage
\clearpage
\FloatBarrier
\appendix
\section*{Appendix}
This appendix provides prompt templates, implementation details, supporting tables, and additional figures referenced in the main text.

\section{Prompt Templates}
\label{app:prompts}

\subsection{Multiple-choice answer prompt}
\label{app:mc_prompt}

For likelihood-based answer selection, we score each answer option independently under the following prompt template:
\begin{quote}
\small\ttfamily\raggedright
Question:\\
<question>\\[2mm]
Choices:\\
0. <choice 0>\\
1. <choice 1>\\
\(\cdots\)\\[2mm]
Answer: \par
\end{quote}

Each candidate answer is appended to the prompt and scored autoregressively. Predictions are then formed using either length-normalized option log-likelihood (\textbf{LL-AVG}) or the unnormalized sum of option log-likelihoods (\textbf{LL-SUM}).

\subsection{Default self-verification prompt}
\label{app:sv_prompt_default}

The default same-model self-verification prompt is:
\begin{quote}
\small\ttfamily\raggedright
You are evaluating whether a proposed answer to a multiple-choice question is correct.\\
Question: <question>\\
Proposed answer: <predicted answer>\\[2mm]
Is the proposed answer correct? Respond with exactly one token: True or False.\\
Answer: \par
\end{quote}

The prompt ends with a trailing space after \texttt{Answer: } to stabilize next-token tokenization.

\subsection{Audit-style self-verification prompt}
\label{app:sv_prompt_audit}

The audit-style prompt used in the prompt-ablation experiments is:
\begin{quote}
\small\ttfamily\raggedright
You are an answer auditor. Determine whether the proposed answer is actually supported by the question.\\
Question: <question>\\
Proposed answer: <predicted answer>\\[2mm]
Is the proposed answer correct? Respond with exactly one token: True or False.\\
Answer: \par
\end{quote}

The audit-style prompt differs from the default prompt only in its opening framing sentence; the question, proposed-answer, and \texttt{True}/\texttt{False} response format are otherwise unchanged.

\section{Additional Experimental Details}
\label{app:details}

\subsection{Datasets, splits, and revisions}
\label{app:datasets}

We evaluate two multiple-choice benchmarks:
\begin{itemize}
    \item \textbf{TruthfulQA-MC}, loaded from \texttt{EleutherAI/truthful\_qa\_mc} using the validation split and the parquet-converted revision \nolinkurl{refs/convert/parquet}.
    \item \textbf{ARC-Challenge}, loaded from \texttt{allenai/ai2\_arc} using the explicit \texttt{ARC-Challenge} configuration and the test split.
\end{itemize}

For ARC, we enforce strict dataset identity and do not permit fallback to the default ARC configuration, which can otherwise mix ARC-Easy and ARC-Challenge examples. Examples whose gold answers cannot be mapped to a valid option index are discarded; in the final reported runs, we did not observe such exclusions.

\subsection{Models}
\label{app:models}

The evaluated model checkpoints are:
\begin{itemize}
    \item \texttt{microsoft/phi-2}
    \item \texttt{Qwen/Qwen2.5-1.5B-Instruct}
    \item \texttt{Qwen/Qwen2.5-7B-Instruct}
    \item \texttt{TinyLlama/TinyLlama-1.1B-Chat-v1.0}
    \item \texttt{deepseek-ai/DeepSeek-R1-Distill-Llama-8B}
\end{itemize}

\subsection{Inference, batching, and quantization}
\label{app:impl}

All experiments use score-based inference without sampling. We fix the random seed to \(42\) and preserve a shuffled evaluation order for reproducibility. The evaluation pipeline is batched and resumable, with outputs checkpointed every 100 examples. We use GPU batch size 8 for both multiple-choice likelihood scoring and self-verification, with maximum sequence length capped at 256 tokens in both stages.

Models are loaded in \texttt{float16} by default. For larger checkpoints, the implementation enables 4-bit NF4 quantization via BitsAndBytes when CUDA is available, with double quantization and \texttt{float16}/\texttt{bfloat16} compute as appropriate. Tokenizer pad tokens are patched to EOS when missing. For \texttt{Phi-2}, we explicitly patch \texttt{pad\_token\_id} in the model configuration before loading to avoid configuration mismatches.

\subsection{True/False token handling}
\label{app:true_false_tokens}

Self-verification confidence is computed from next-token logits rather than from free-text generation. To make this robust across tokenizers, we aggregate probability mass over common single-token variants of \texttt{True} and \texttt{False}, including leading-space and uppercase forms. If no usable True/False tokenization is available, the implementation falls back to \(\{1,0\}\) token variants as a safeguard.

\section{Additional Results}
\label{app:results}

\subsection{Calibration results}
\label{app:calibration}

Table~\ref{tab:appendix_calibration} reports additional calibration metrics for the main confidence signals, including Brier score and expected calibration error with 10 bins (ECE-10).

\begin{table}[!t]
\centering
\scriptsize
\setlength{\tabcolsep}{3.5pt}
\renewcommand{\arraystretch}{0.95}
\caption{Additional calibration metrics for the main confidence signals, including Brier score and expected calibration error with 10 bins (ECE-10).}
\label{tab:appendix_calibration}
\resizebox{\textwidth}{!}{%
\begin{tabular}{lllrrrrrr}
\toprule
Dataset & Model & Prompt & Brier (LL-AVG) & Brier (Self-Verify) & Brier (LL-SUM) & ECE10 (LL-AVG) & ECE10 (Self-Verify) & ECE10 (LL-SUM) \\
\midrule
ARC-Challenge & DeepSeek-R1-Distill-8B & audit\_v1 & 0.427 & 0.360 & 0.582 & 0.383 & 0.320 & 0.576 \\
ARC-Challenge & DeepSeek-R1-Distill-8B & default & 0.427 & 0.370 & 0.582 & 0.383 & 0.339 & 0.576 \\
ARC-Challenge & Phi-2 & audit\_v1 & 0.269 & 0.210 & 0.228 & 0.134 & 0.080 & 0.131 \\
ARC-Challenge & Phi-2 & default & 0.269 & 0.204 & 0.228 & 0.134 & 0.064 & 0.131 \\
ARC-Challenge & Qwen-1.5B & audit\_v1 & 0.272 & 0.289 & 0.268 & 0.119 & 0.275 & 0.216 \\
ARC-Challenge & Qwen-1.5B & default & 0.272 & 0.287 & 0.268 & 0.119 & 0.272 & 0.216 \\
ARC-Challenge & Qwen-7B & audit\_v1 & 0.268 & 0.253 & 0.204 & 0.166 & 0.252 & 0.155 \\
ARC-Challenge & Qwen-7B & default & 0.268 & 0.238 & 0.204 & 0.166 & 0.239 & 0.155 \\
ARC-Challenge & TinyLlama-1.1B & audit\_v1 & 0.221 & 0.453 & 0.334 & 0.080 & 0.502 & 0.336 \\
ARC-Challenge & TinyLlama-1.1B & default & 0.221 & 0.492 & 0.334 & 0.080 & 0.538 & 0.336 \\
TruthfulQA-MC & DeepSeek-R1-Distill-8B & audit\_v1 & 0.419 & 0.390 & 0.550 & 0.399 & 0.348 & 0.556 \\
TruthfulQA-MC & DeepSeek-R1-Distill-8B & default & 0.419 & 0.392 & 0.550 & 0.399 & 0.340 & 0.556 \\
TruthfulQA-MC & Phi-2 & audit\_v1 & 0.255 & 0.241 & 0.289 & 0.109 & 0.067 & 0.237 \\
TruthfulQA-MC & Phi-2 & default & 0.255 & 0.242 & 0.289 & 0.109 & 0.073 & 0.237 \\
TruthfulQA-MC & Qwen-1.5B & audit\_v1 & 0.244 & 0.316 & 0.356 & 0.075 & 0.267 & 0.351 \\
TruthfulQA-MC & Qwen-1.5B & default & 0.244 & 0.355 & 0.356 & 0.075 & 0.307 & 0.351 \\
TruthfulQA-MC & Qwen-7B & audit\_v1 & 0.260 & 0.372 & 0.271 & 0.123 & 0.361 & 0.252 \\
TruthfulQA-MC & Qwen-7B & default & 0.260 & 0.363 & 0.271 & 0.123 & 0.351 & 0.252 \\
TruthfulQA-MC & TinyLlama-1.1B & audit\_v1 & 0.194 & 0.488 & 0.344 & 0.040 & 0.532 & 0.392 \\
TruthfulQA-MC & TinyLlama-1.1B & default & 0.194 & 0.552 & 0.344 & 0.040 & 0.591 & 0.392 \\
\bottomrule
\end{tabular}%
}
\end{table}
\FloatBarrier

\subsection{Auxiliary baseline comparisons}
\label{app:aux_baselines}

To contextualize the main comparison, we also compute several auxiliary confidence baselines from the multiple-choice answer distribution:
\begin{itemize}
    \item \textbf{Margin}: the gap between the top two LL-AVG option probabilities;
    \item \textbf{EntropyConf}: one minus normalized predictive entropy;
    \item \textbf{LL-AVG-T}: a temperature-scaled version of LL-AVG fit on a held-out 20\% calibration subset, with a minimum of 50 calibration examples.
\end{itemize}

These baselines do not alter the answer prediction pipeline used in the main experiments; they are included only as supporting comparisons. Table~\ref{tab:appendix_aux_baselines} reports their results. They do not overturn the main story: some are competitive in specific settings, but none consistently dominates the stronger baselines used in the main text.

\begin{table}[!t]
\centering
\scriptsize
\setlength{\tabcolsep}{3pt}
\renewcommand{\arraystretch}{0.95}
\caption{Auxiliary confidence baselines computed from the multiple-choice answer distribution. Margin denotes the gap between the top two LL-AVG option probabilities; EntropyConf denotes one minus normalized predictive entropy; LL-AVG-T denotes temperature-scaled LL-AVG fit on a held-out 20\% calibration split.}
\label{tab:appendix_aux_baselines}
\resizebox{\textwidth}{!}{%
\begin{tabular}{lllrrrrrrrrrrr}
\toprule
Dataset & Model & Prompt & AUROC (LL-AVG) & AUROC (SV) & AUROC (Margin) & AUROC (EntropyConf) & AUROC (LL-AVG-T) & AURC (LL-AVG) & AURC (SV) & AURC (Margin) & AURC (EntropyConf) & AURC (LL-AVG-T) \\
\midrule
ARC-Challenge & Qwen-1.5B & audit\_v1 & 0.557 & 0.770 & 0.578 & 0.531 & 0.545 & 0.471 & 0.295 & 0.459 & 0.488 & 0.483 \\
ARC-Challenge & Qwen-1.5B & default & 0.557 & 0.765 & 0.578 & 0.531 & 0.545 & 0.471 & 0.298 & 0.459 & 0.488 & 0.483 \\
ARC-Challenge & Qwen-7B & audit\_v1 & 0.555 & 0.879 & 0.585 & 0.515 & 0.550 & 0.363 & 0.148 & 0.348 & 0.384 & 0.367 \\
ARC-Challenge & Qwen-7B & default & 0.555 & 0.886 & 0.585 & 0.515 & 0.550 & 0.363 & 0.143 & 0.348 & 0.384 & 0.367 \\
ARC-Challenge & TinyLlama-1.1B & audit\_v1 & 0.484 & 0.525 & 0.505 & 0.471 & 0.481 & 0.731 & 0.702 & 0.722 & 0.735 & 0.734 \\
ARC-Challenge & TinyLlama-1.1B & default & 0.484 & 0.525 & 0.505 & 0.471 & 0.481 & 0.731 & 0.700 & 0.722 & 0.735 & 0.734 \\
ARC-Challenge & DeepSeek-R1-Distill-8B & audit\_v1 & 0.511 & 0.489 & 0.514 & 0.514 & 0.513 & 0.721 & 0.737 & 0.720 & 0.719 & 0.721 \\
ARC-Challenge & DeepSeek-R1-Distill-8B & default & 0.511 & 0.463 & 0.514 & 0.514 & 0.513 & 0.721 & 0.751 & 0.720 & 0.719 & 0.721 \\
ARC-Challenge & Phi-2 & audit\_v1 & 0.547 & 0.755 & 0.573 & 0.525 & 0.543 & 0.427 & 0.278 & 0.414 & 0.441 & 0.431 \\
ARC-Challenge & Phi-2 & default & 0.547 & 0.759 & 0.573 & 0.525 & 0.543 & 0.427 & 0.274 & 0.414 & 0.441 & 0.431 \\
TruthfulQA-MC & Qwen-1.5B & audit\_v1 & 0.611 & 0.620 & 0.643 & 0.554 & 0.586 & 0.487 & 0.499 & 0.464 & 0.527 & 0.510 \\
TruthfulQA-MC & Qwen-1.5B & default & 0.611 & 0.548 & 0.643 & 0.554 & 0.586 & 0.487 & 0.541 & 0.464 & 0.527 & 0.510 \\
TruthfulQA-MC & Qwen-7B & audit\_v1 & 0.593 & 0.665 & 0.640 & 0.531 & 0.576 & 0.415 & 0.366 & 0.386 & 0.460 & 0.432 \\
TruthfulQA-MC & Qwen-7B & default & 0.593 & 0.667 & 0.640 & 0.531 & 0.576 & 0.415 & 0.370 & 0.386 & 0.460 & 0.432 \\
TruthfulQA-MC & TinyLlama-1.1B & audit\_v1 & 0.562 & 0.372 & 0.518 & 0.574 & 0.563 & 0.690 & 0.798 & 0.710 & 0.691 & 0.691 \\
TruthfulQA-MC & TinyLlama-1.1B & default & 0.562 & 0.363 & 0.518 & 0.574 & 0.563 & 0.690 & 0.806 & 0.710 & 0.691 & 0.691 \\
TruthfulQA-MC & DeepSeek-R1-Distill-8B & audit\_v1 & 0.614 & 0.524 & 0.617 & 0.612 & 0.614 & 0.513 & 0.596 & 0.513 & 0.514 & 0.513 \\
TruthfulQA-MC & DeepSeek-R1-Distill-8B & default & 0.614 & 0.494 & 0.617 & 0.612 & 0.614 & 0.513 & 0.604 & 0.513 & 0.514 & 0.513 \\
TruthfulQA-MC & Phi-2 & audit\_v1 & 0.590 & 0.624 & 0.625 & 0.542 & 0.585 & 0.478 & 0.460 & 0.452 & 0.515 & 0.483 \\
TruthfulQA-MC & Phi-2 & default & 0.590 & 0.608 & 0.625 & 0.542 & 0.585 & 0.478 & 0.465 & 0.452 & 0.515 & 0.483 \\
\bottomrule
\end{tabular}
}
\end{table}
\FloatBarrier

\subsection{Expanded operating-point results}
\label{app:operating_points}

The main paper reports representative selective-prediction operating points. For completeness, Table~\ref{tab:operating_points_appendix} reports the full set of operating-point comparisons used in the analysis.

\begin{table}[!t]
\centering
\scriptsize
\setlength{\tabcolsep}{3pt}
\renewcommand{\arraystretch}{0.95}
\caption{Full operating-point comparisons used in the selective-prediction analysis. Lower error at fixed coverage and higher coverage at fixed error indicate better abstention behavior.}
\label{tab:operating_points_appendix}
\resizebox{\textwidth}{!}{%
\begin{tabular}{llllrrrr}
\toprule
Dataset & Model & Prompt & Signal & err@80\%cov & err@50\%cov & cov@20\%err & cov@10\%err \\
\midrule
ARC-Challenge & DeepSeek-R1-Distill-8B & audit\_v1 & LL-AVG      & 0.720 & 0.720 & 0.000 & 0.000 \\
ARC-Challenge & DeepSeek-R1-Distill-8B & audit\_v1 & LL-SUM      & 0.737 & 0.727 & 0.000 & 0.000 \\
ARC-Challenge & DeepSeek-R1-Distill-8B & audit\_v1 & Self-Verify & 0.719 & 0.729 & 0.000 & 0.000 \\
ARC-Challenge & DeepSeek-R1-Distill-8B & default   & LL-AVG      & 0.720 & 0.720 & 0.000 & 0.000 \\
ARC-Challenge & DeepSeek-R1-Distill-8B & default   & LL-SUM      & 0.737 & 0.727 & 0.000 & 0.000 \\
ARC-Challenge & DeepSeek-R1-Distill-8B & default   & Self-Verify & 0.726 & 0.744 & 0.001 & 0.001 \\
ARC-Challenge & Phi-2                  & audit\_v1 & LL-AVG      & 0.447 & 0.451 & 0.004 & 0.002 \\
ARC-Challenge & Phi-2                  & audit\_v1 & LL-SUM      & 0.327 & 0.253 & 0.412 & 0.061 \\
ARC-Challenge & Phi-2                  & audit\_v1 & Self-Verify & 0.388 & 0.280 & 0.287 & 0.051 \\
ARC-Challenge & Phi-2                  & default   & LL-AVG      & 0.447 & 0.451 & 0.004 & 0.002 \\
ARC-Challenge & Phi-2                  & default   & LL-SUM      & 0.327 & 0.253 & 0.412 & 0.061 \\
ARC-Challenge & Phi-2                  & default   & Self-Verify & 0.387 & 0.268 & 0.317 & 0.073 \\
ARC-Challenge & Qwen-1.5B              & audit\_v1 & LL-AVG      & 0.485 & 0.462 & 0.007 & 0.002 \\
ARC-Challenge & Qwen-1.5B              & audit\_v1 & LL-SUM      & 0.374 & 0.299 & 0.241 & 0.094 \\
ARC-Challenge & Qwen-1.5B              & audit\_v1 & Self-Verify & 0.427 & 0.311 & 0.286 & 0.111 \\
ARC-Challenge & Qwen-1.5B              & default   & LL-AVG      & 0.485 & 0.462 & 0.007 & 0.002 \\
ARC-Challenge & Qwen-1.5B              & default   & LL-SUM      & 0.374 & 0.299 & 0.241 & 0.094 \\
ARC-Challenge & Qwen-1.5B              & default   & Self-Verify & 0.431 & 0.321 & 0.262 & 0.096 \\
ARC-Challenge & Qwen-7B                & audit\_v1 & LL-AVG      & 0.369 & 0.362 & 0.026 & 0.009 \\
ARC-Challenge & Qwen-7B                & audit\_v1 & LL-SUM      & 0.235 & 0.148 & 0.679 & 0.334 \\
ARC-Challenge & Qwen-7B                & audit\_v1 & Self-Verify & 0.273 & 0.121 & 0.659 & 0.454 \\
ARC-Challenge & Qwen-7B                & default   & LL-AVG      & 0.369 & 0.362 & 0.026 & 0.009 \\
ARC-Challenge & Qwen-7B                & default   & LL-SUM      & 0.235 & 0.148 & 0.679 & 0.334 \\
ARC-Challenge & Qwen-7B                & default   & Self-Verify & 0.271 & 0.121 & 0.666 & 0.463 \\
ARC-Challenge & TinyLlama-1.1B         & audit\_v1 & LL-AVG      & 0.721 & 0.729 & 0.001 & 0.001 \\
ARC-Challenge & TinyLlama-1.1B         & audit\_v1 & LL-SUM      & 0.717 & 0.706 & 0.000 & 0.000 \\
ARC-Challenge & TinyLlama-1.1B         & audit\_v1 & Self-Verify & 0.718 & 0.705 & 0.001 & 0.001 \\
ARC-Challenge & TinyLlama-1.1B         & default   & LL-AVG      & 0.721 & 0.729 & 0.001 & 0.001 \\
ARC-Challenge & TinyLlama-1.1B         & default   & LL-SUM      & 0.717 & 0.706 & 0.000 & 0.000 \\
ARC-Challenge & TinyLlama-1.1B         & default   & Self-Verify & 0.710 & 0.708 & 0.001 & 0.001 \\
TruthfulQA-MC & DeepSeek-R1-Distill-8B & audit\_v1 & LL-AVG      & 0.565 & 0.512 & 0.015 & 0.007 \\
TruthfulQA-MC & DeepSeek-R1-Distill-8B & audit\_v1 & LL-SUM      & 0.601 & 0.558 & 0.001 & 0.001 \\
TruthfulQA-MC & DeepSeek-R1-Distill-8B & audit\_v1 & Self-Verify & 0.583 & 0.591 & 0.000 & 0.000 \\
TruthfulQA-MC & DeepSeek-R1-Distill-8B & default   & LL-AVG      & 0.565 & 0.512 & 0.015 & 0.007 \\
TruthfulQA-MC & DeepSeek-R1-Distill-8B & default   & LL-SUM      & 0.601 & 0.558 & 0.001 & 0.001 \\
TruthfulQA-MC & DeepSeek-R1-Distill-8B & default   & Self-Verify & 0.592 & 0.608 & 0.003 & 0.003 \\
TruthfulQA-MC & Phi-2                  & audit\_v1 & LL-AVG      & 0.501 & 0.456 & 0.009 & 0.006 \\
TruthfulQA-MC & Phi-2                  & audit\_v1 & LL-SUM      & 0.521 & 0.442 & 0.126 & 0.019 \\
TruthfulQA-MC & Phi-2                  & audit\_v1 & Self-Verify & 0.484 & 0.436 & 0.007 & 0.001 \\
TruthfulQA-MC & Phi-2                  & default   & LL-AVG      & 0.501 & 0.456 & 0.009 & 0.006 \\
TruthfulQA-MC & Phi-2                  & default   & LL-SUM      & 0.521 & 0.442 & 0.126 & 0.019 \\
TruthfulQA-MC & Phi-2                  & default   & Self-Verify & 0.494 & 0.430 & 0.012 & 0.007 \\
TruthfulQA-MC & Qwen-1.5B              & audit\_v1 & LL-AVG      & 0.528 & 0.491 & 0.000 & 0.000 \\
TruthfulQA-MC & Qwen-1.5B              & audit\_v1 & LL-SUM      & 0.589 & 0.538 & 0.010 & 0.003 \\
TruthfulQA-MC & Qwen-1.5B              & audit\_v1 & Self-Verify & 0.508 & 0.462 & 0.000 & 0.000 \\
TruthfulQA-MC & Qwen-1.5B              & default   & LL-AVG      & 0.528 & 0.491 & 0.000 & 0.000 \\
TruthfulQA-MC & Qwen-1.5B              & default   & LL-SUM      & 0.589 & 0.538 & 0.010 & 0.003 \\
TruthfulQA-MC & Qwen-1.5B              & default   & Self-Verify & 0.545 & 0.523 & 0.000 & 0.000 \\
TruthfulQA-MC & Qwen-7B                & audit\_v1 & LL-AVG      & 0.444 & 0.392 & 0.007 & 0.004 \\
TruthfulQA-MC & Qwen-7B                & audit\_v1 & LL-SUM      & 0.375 & 0.257 & 0.351 & 0.118 \\
TruthfulQA-MC & Qwen-7B                & audit\_v1 & Self-Verify & 0.404 & 0.354 & 0.007 & 0.004 \\
TruthfulQA-MC & Qwen-7B                & default   & LL-AVG      & 0.444 & 0.392 & 0.007 & 0.004 \\
TruthfulQA-MC & Qwen-7B                & default   & LL-SUM      & 0.375 & 0.257 & 0.351 & 0.118 \\
TruthfulQA-MC & Qwen-7B                & default   & Self-Verify & 0.395 & 0.351 & 0.009 & 0.003 \\
TruthfulQA-MC & TinyLlama-1.1B         & audit\_v1 & LL-AVG      & 0.722 & 0.702 & 0.001 & 0.001 \\
TruthfulQA-MC & TinyLlama-1.1B         & audit\_v1 & LL-SUM      & 0.797 & 0.792 & 0.000 & 0.000 \\
TruthfulQA-MC & TinyLlama-1.1B         & audit\_v1 & Self-Verify & 0.779 & 0.804 & 0.000 & 0.000 \\
TruthfulQA-MC & TinyLlama-1.1B         & default   & LL-AVG      & 0.722 & 0.702 & 0.001 & 0.001 \\
TruthfulQA-MC & TinyLlama-1.1B         & default   & LL-SUM      & 0.797 & 0.792 & 0.000 & 0.000 \\
TruthfulQA-MC & TinyLlama-1.1B         & default   & Self-Verify & 0.775 & 0.816 & 0.000 & 0.000 \\
\bottomrule
\end{tabular}%
}
\end{table}
\FloatBarrier

\subsection{Prompt-ablation details}
\label{app:prompt_ablation}

Table~\ref{tab:prompt_ablation_appendix} reports the full prompt-ablation results across datasets, models, and verification prompt variants. The main text emphasizes the qualitative difference between ARC-Challenge, where prompt sensitivity is generally limited, and TruthfulQA-MC, where some smaller models exhibit substantially greater sensitivity to verification prompt wording.

\begin{table}[!t]
\centering
\small
\setlength{\tabcolsep}{4pt}
\renewcommand{\arraystretch}{0.95}
\caption{Prompt-ablation results for the Self-Verify confidence signal across datasets, models, and verification prompt variants.}
\label{tab:prompt_ablation_appendix}
\resizebox{\textwidth}{!}{%
\begin{tabular}{llrrrrrr}
\toprule
Dataset & Model & AURC (audit\_v1) & AURC (default) & AUROC (audit\_v1) & AUROC (default) & $|\Delta \mathrm{AUROC}|$ & $|\Delta \mathrm{AURC}|$ \\
\midrule
ARC-Challenge & DeepSeek-R1-Distill-8B & 0.738 & 0.751 & 0.489 & 0.463 & 0.026 & 0.013 \\
ARC-Challenge & Phi-2 & 0.278 & 0.275 & 0.755 & 0.759 & 0.004 & 0.003 \\
ARC-Challenge & Qwen-1.5B & 0.296 & 0.298 & 0.770 & 0.765 & 0.005 & 0.003 \\
ARC-Challenge & Qwen-7B & 0.149 & 0.143 & 0.879 & 0.886 & 0.008 & 0.005 \\
ARC-Challenge & TinyLlama-1.1B & 0.702 & 0.700 & 0.525 & 0.525 & 0.000 & 0.002 \\
TruthfulQA-MC & DeepSeek-R1-Distill-8B & 0.597 & 0.605 & 0.524 & 0.494 & 0.030 & 0.007 \\
TruthfulQA-MC & Phi-2 & 0.461 & 0.465 & 0.624 & 0.608 & 0.016 & 0.005 \\
TruthfulQA-MC & Qwen-1.5B & 0.500 & 0.543 & 0.620 & 0.548 & 0.073 & 0.043 \\
TruthfulQA-MC & Qwen-7B & 0.367 & 0.371 & 0.665 & 0.667 & 0.003 & 0.004 \\
TruthfulQA-MC & TinyLlama-1.1B & 0.800 & 0.807 & 0.372 & 0.363 & 0.009 & 0.008 \\
\bottomrule
\end{tabular}
}
\end{table}
\FloatBarrier

\subsection{Bootstrap confidence intervals}
\label{app:bootstrap}

To assess whether self-verification meaningfully changes correctness ranking relative to LL-AVG, we estimate bootstrap confidence intervals for
\[
\Delta \mathrm{AUROC}
=
\mathrm{AUROC}(c_{\mathrm{SV}}, y_{\mathrm{avg}})
-
\mathrm{AUROC}(c_{\mathrm{LL\mbox{-}AVG}}, y_{\mathrm{avg}}).
\]
For each model--dataset pair, we use 2000 bootstrap resamples with seed \(42\), discarding replicates that contain only one class. Because we evaluate multiple models and datasets, these intervals are not corrected for multiple comparisons and should be interpreted as exploratory. Full results are reported in Table~\ref{tab:delta_auroc} in the main text.

\FloatBarrier

\section{Additional Figures}
\label{app:figures}

\subsection{AUROC by dataset and prompt}
\label{app:fig_auroc}

\begin{figure}[t]
    \centering
    \includegraphics[width=0.85\linewidth]{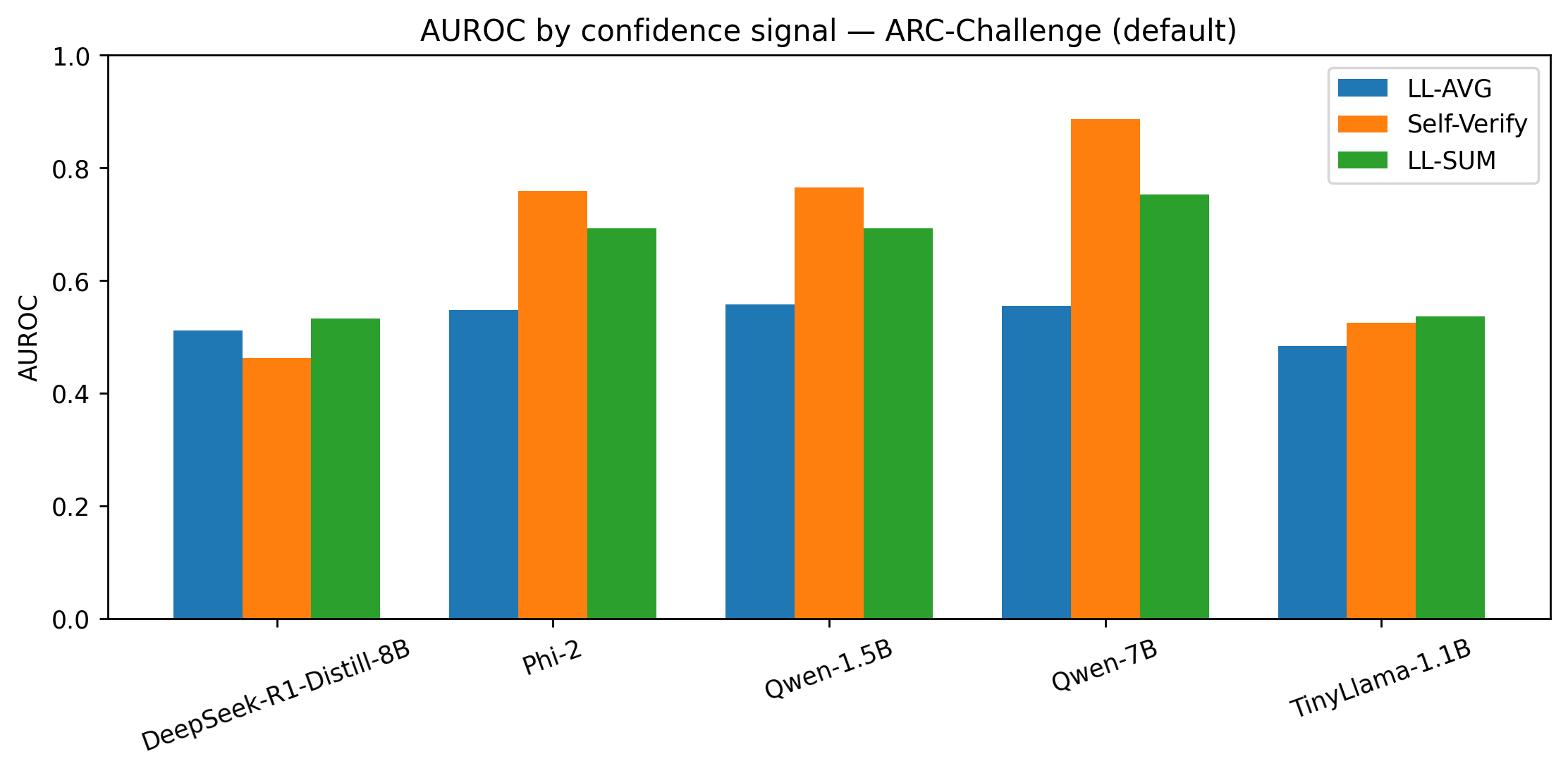}
    \caption{AUROC comparison on ARC-Challenge under the default verification prompt.}
    \label{fig:app_auroc_arc_default}
\end{figure}

\begin{figure}[t]
    \centering
    \includegraphics[width=0.85\linewidth]{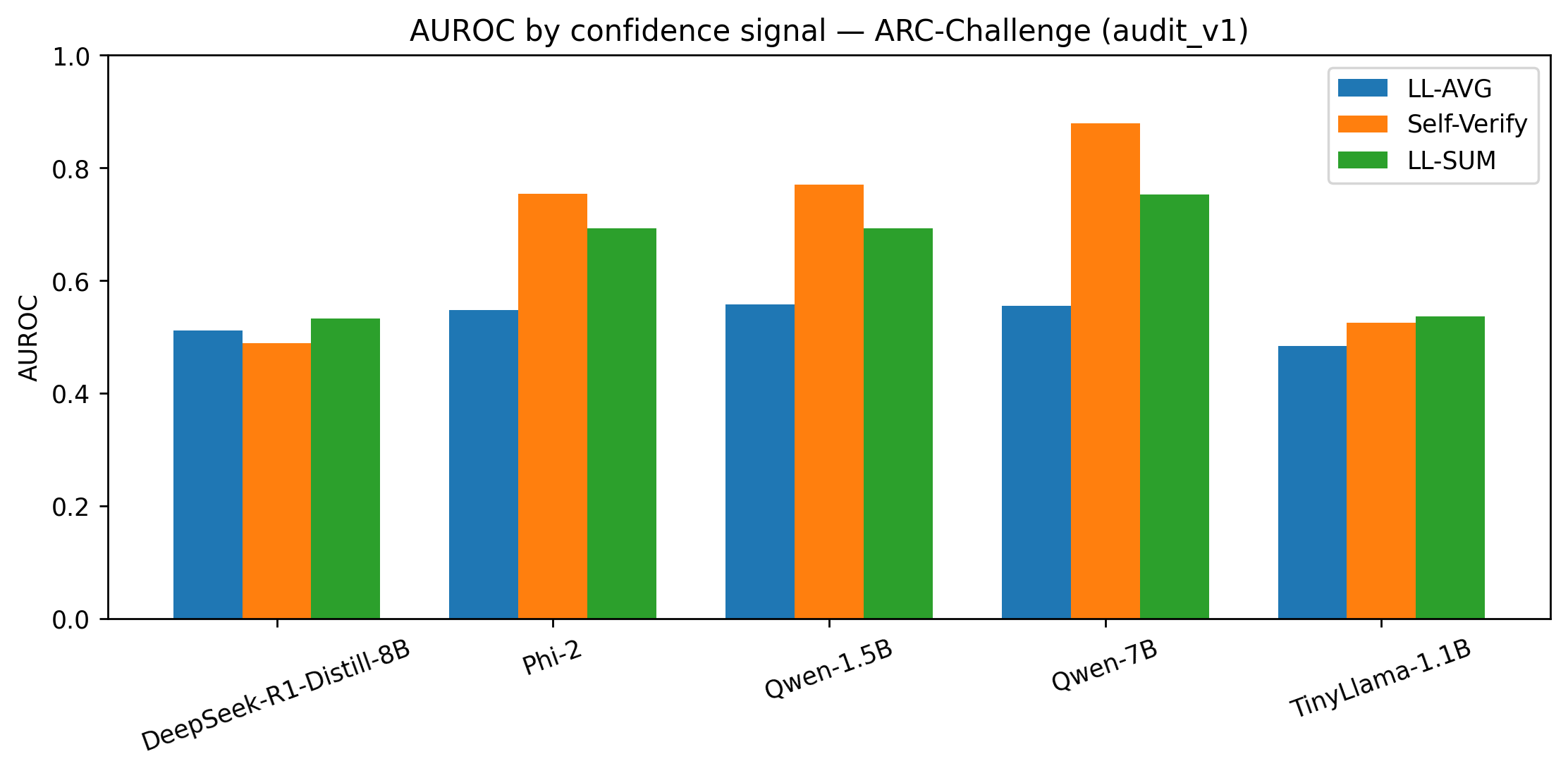}
    \caption{AUROC comparison on ARC-Challenge under the audit-style verification prompt.}
    \label{fig:app_auroc_arc_audit}
\end{figure}

\begin{figure}[t]
    \centering
    \includegraphics[width=0.85\linewidth]{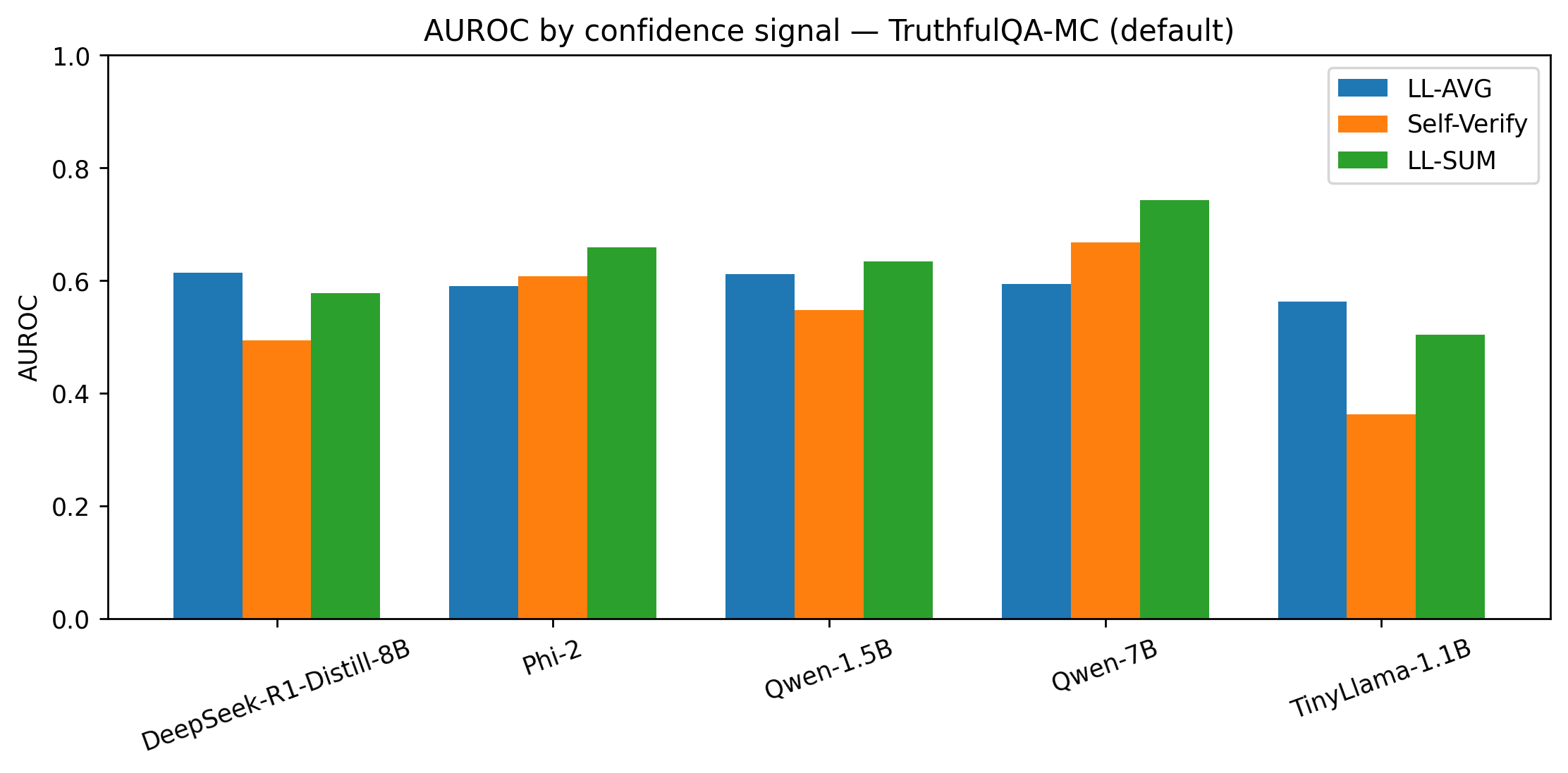}
    \caption{AUROC comparison on TruthfulQA-MC under the default verification prompt.}
    \label{fig:app_auroc_truthful_default}
\end{figure}

\begin{figure}[t]
    \centering
    \includegraphics[width=0.85\linewidth]{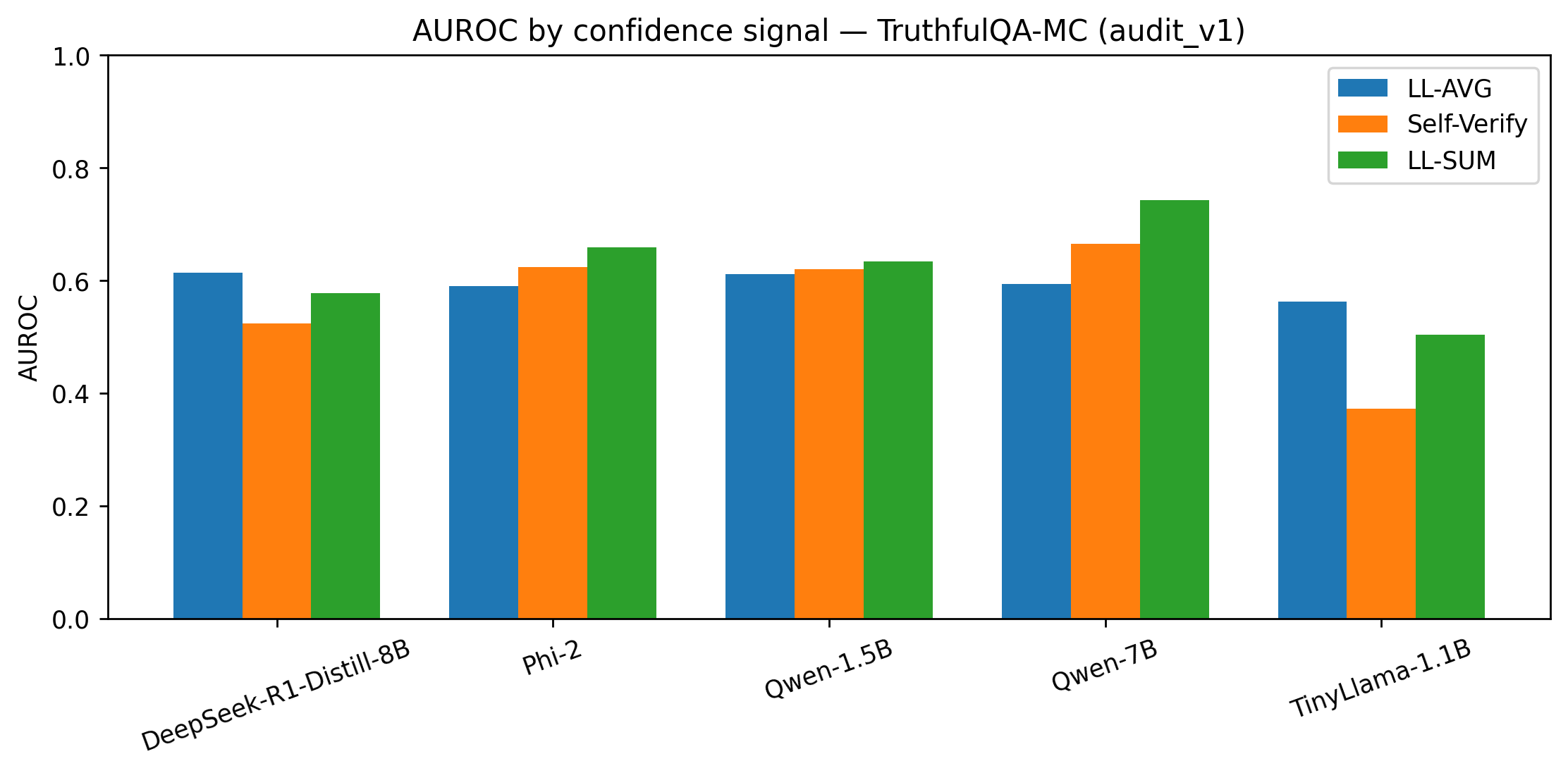}
    \caption{AUROC comparison on TruthfulQA-MC under the audit-style verification prompt.}
    \label{fig:app_auroc_truthful_audit}
\end{figure}

Figures~\ref{fig:app_auroc_arc_default}--\ref{fig:app_auroc_truthful_audit} report the full AUROC comparisons across datasets and verification prompt variants.

\FloatBarrier

\subsection{AURC by dataset and prompt}
\label{app:fig_aurc}

\begin{figure}[t]
    \centering
    \includegraphics[width=0.85\linewidth]{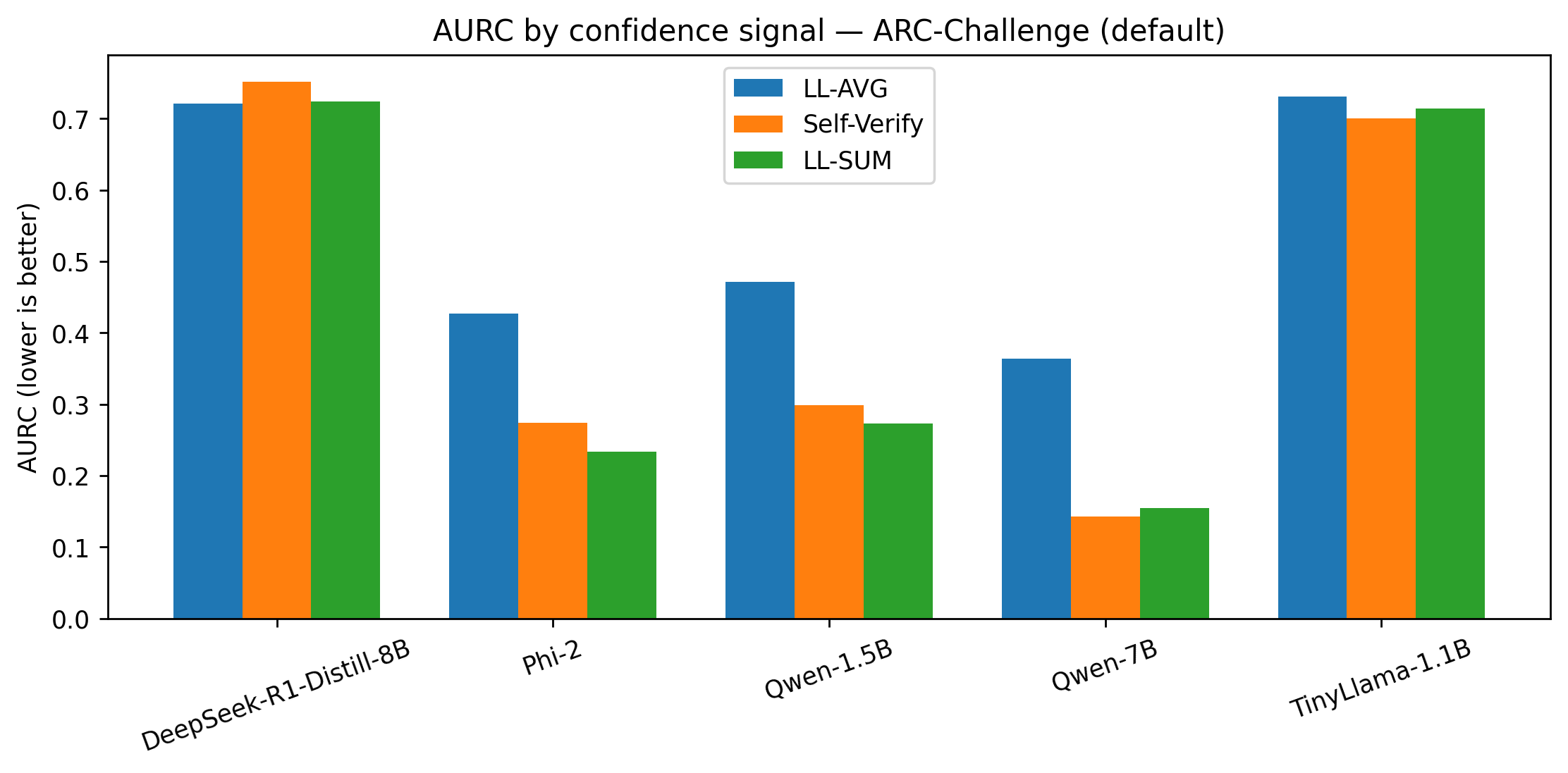}
    \caption{AURC comparison on ARC-Challenge under the default verification prompt.}
    \label{fig:app_aurc_arc_default}
\end{figure}

\begin{figure}[t]
    \centering
    \includegraphics[width=0.85\linewidth]{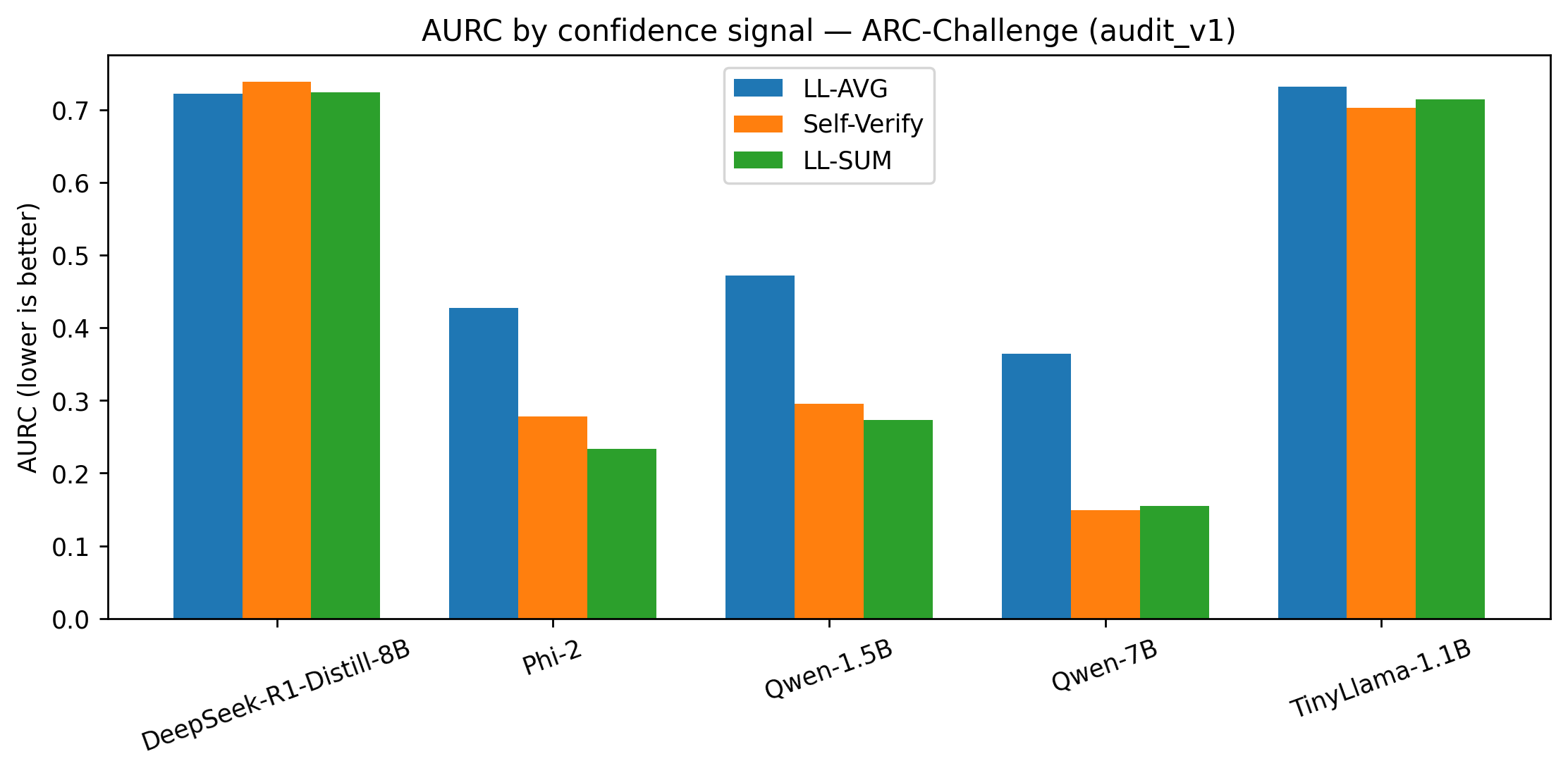}
    \caption{AURC comparison on ARC-Challenge under the audit-style verification prompt.}
    \label{fig:app_aurc_arc_audit}
\end{figure}

\begin{figure}[t]
    \centering
    \includegraphics[width=0.85\linewidth]{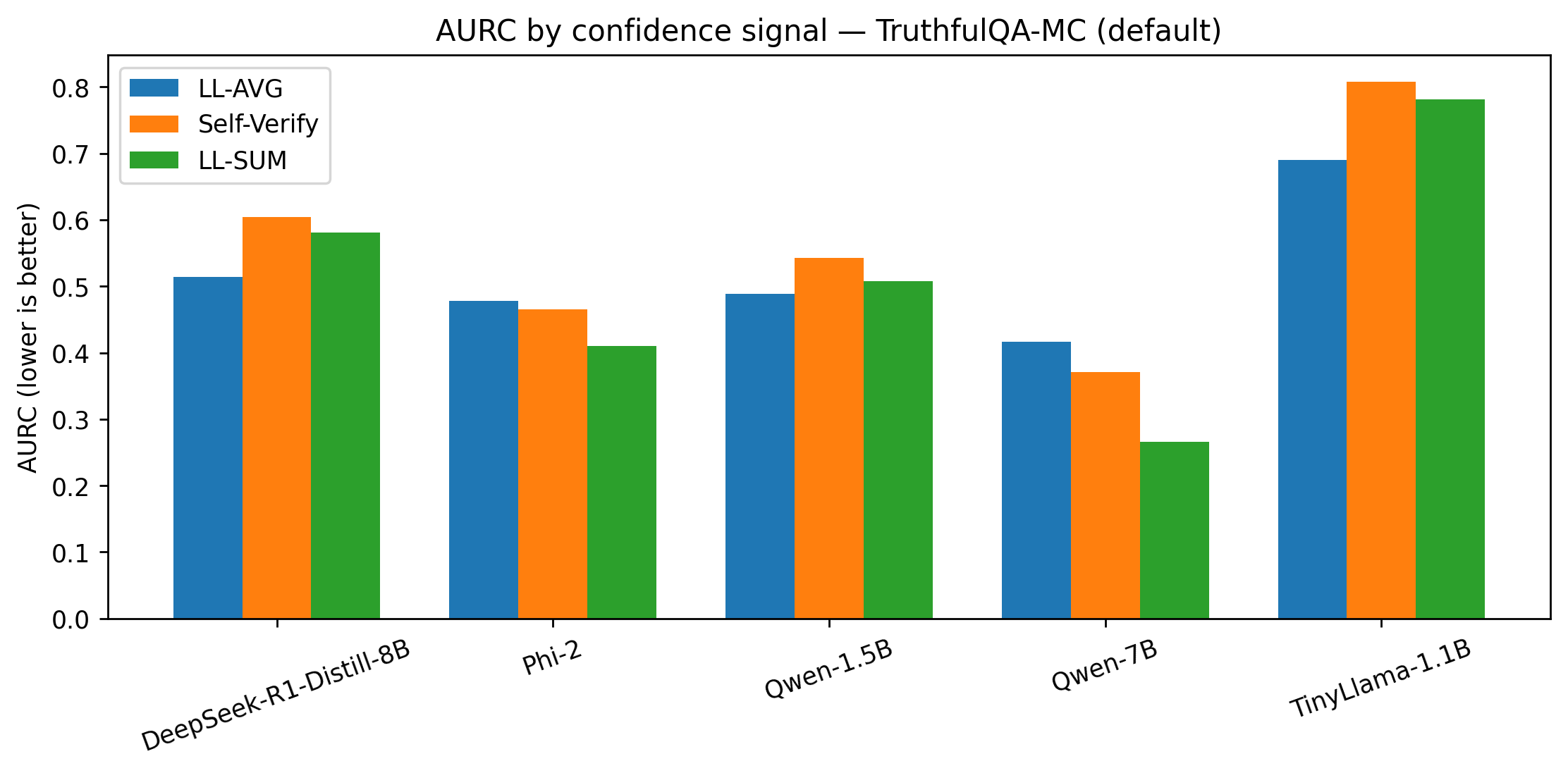}
    \caption{AURC comparison on TruthfulQA-MC under the default verification prompt.}
    \label{fig:app_aurc_truthful_default}
\end{figure}

\begin{figure}[t]
    \centering
    \includegraphics[width=0.85\linewidth]{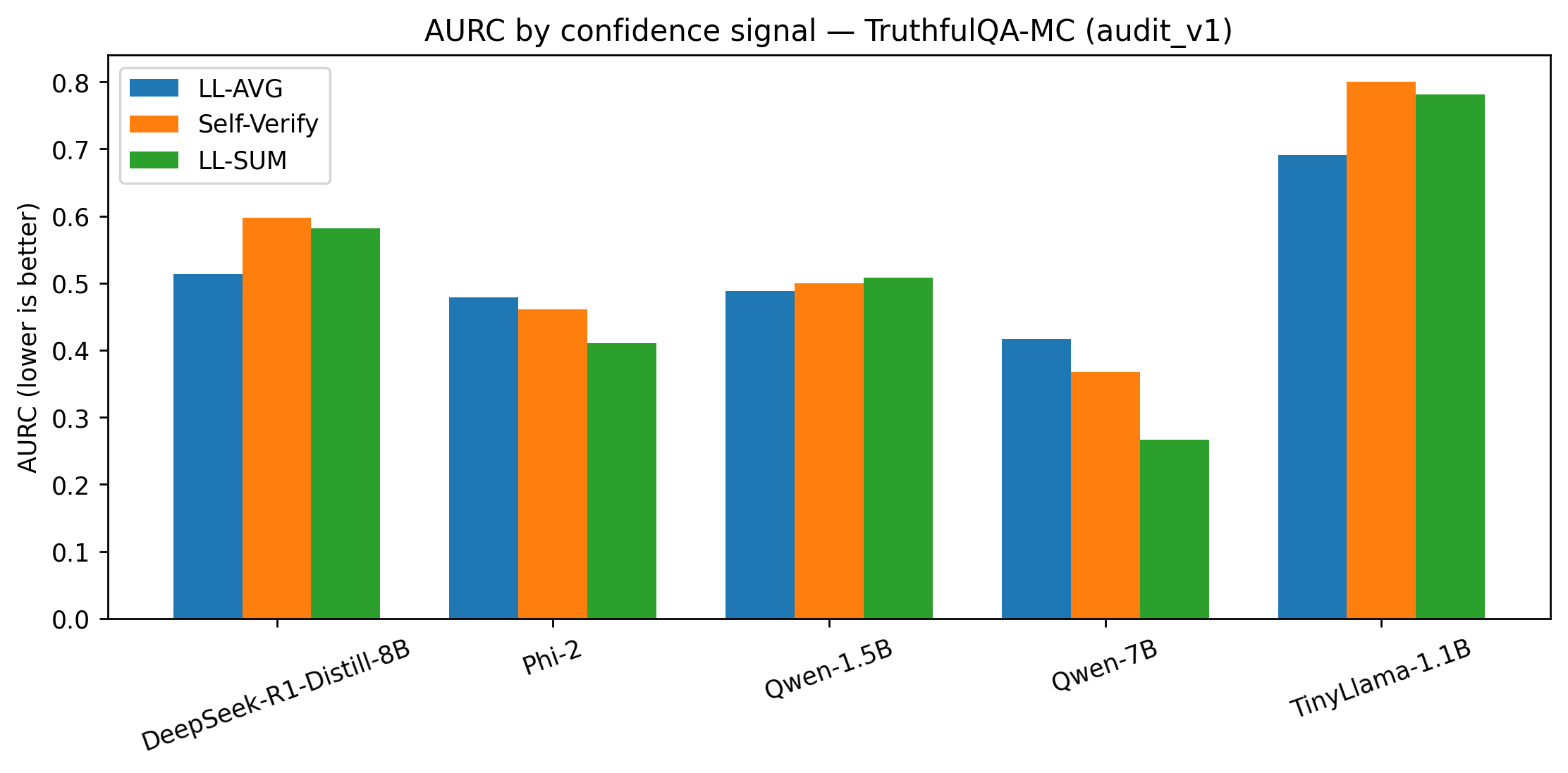}
    \caption{AURC comparison on TruthfulQA-MC under the audit-style verification prompt.}
    \label{fig:app_aurc_truthful_audit}
\end{figure}

Figures~\ref{fig:app_aurc_arc_default}--\ref{fig:app_aurc_truthful_audit} report the corresponding AURC comparisons.

\FloatBarrier

\subsection{Representative risk--coverage and family-scale figures}
\label{app:fig_rc_scale}

\begin{figure}[t]
    \centering
    \includegraphics[width=0.85\linewidth]{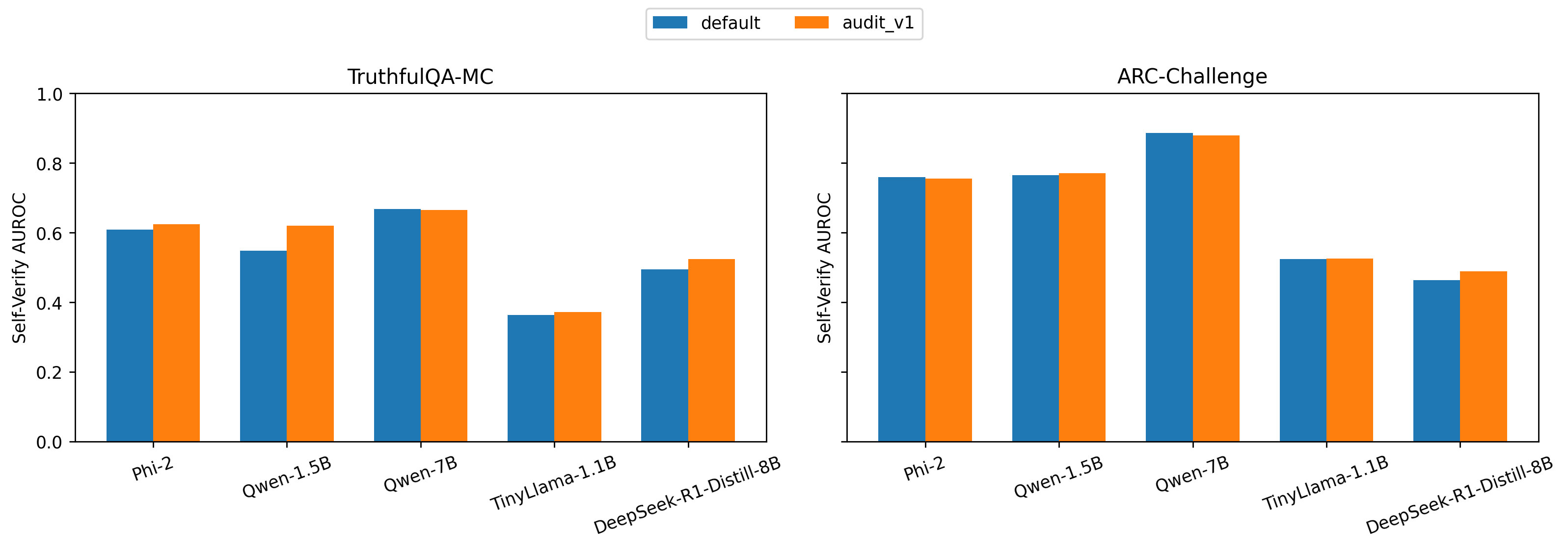}
    \caption{Prompt-ablation AUROC summary across datasets and models.}
    \label{fig:app_prompt_ablation}
\end{figure}

\begin{figure}[t]
    \centering
    \includegraphics[width=0.85\linewidth]{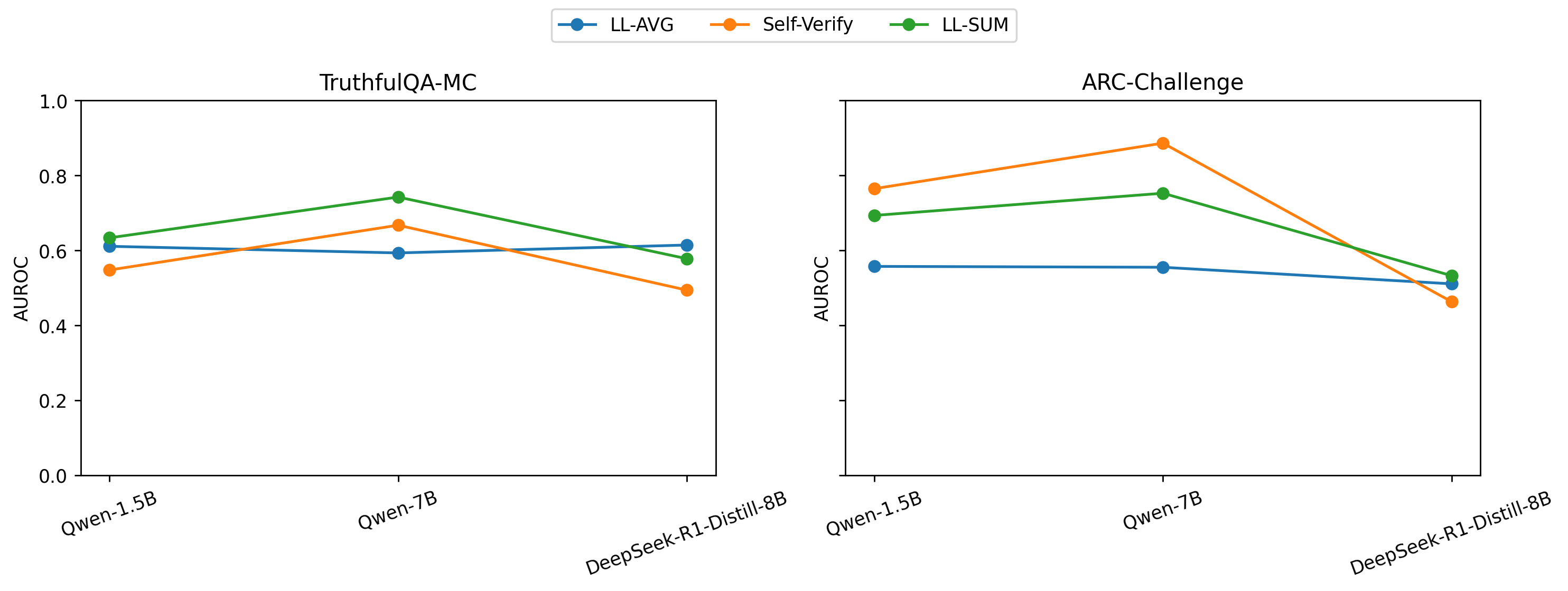}
    \caption{Scale and family effects in self-verification performance.}
    \label{fig:app_scale_family}
\end{figure}

The representative risk--coverage comparison is shown as Figure~\ref{fig:risk_coverage_examples} in the main text, while Figures~\ref{fig:app_prompt_ablation} and \ref{fig:app_scale_family} summarize prompt-ablation and scale/family effects.

\end{document}